\RequirePackage{fix-cm}
\documentclass{svjour3}
\usepackage[top=1.1in, bottom=1.2in, left=1.2in, right=1.2in]{geometry}
\usepackage{graphicx}
\usepackage[numbers]{natbib} 
\usepackage{amsmath}
\usepackage{amstext}
\usepackage{amsfonts}
\usepackage{amssymb}

\usepackage{soul}
\usepackage{multirow}
\usepackage{rotating}
\usepackage[usenames,dvipsnames]{color}
\usepackage{algorithm}
\usepackage[noend]{algpseudocode}
\usepackage{textcomp}
\usepackage{gensymb}
\usepackage{placeins}
\usepackage{caption}
\usepackage{array}

\ifx\mjdraft\undefined
 
 \newcommand{\new}[1]{\textcolor{black}{#1}}
 
\else
 
 \newcommand{\new}[1]{\textcolor{blue}{NEW: #1}}

\fi
\newcommand{\myparagraph}[1]{\noindent \textbf{#1}}
\newcommand{\reduceFigSpace}{\vspace{-0.1cm}} 

\newcommand{\SOTA}{state-of-the-art }
\newcommand{\eg}{\textit{e.g.} }
\newcommand{\ie}{\textit{i.e.} }
\newcommand{\etal}{\textit{et al.} }

\usepackage{hyperref}
\usepackage{enumerate}

\usepackage[switch,columnwise]{lineno}
\begin{document}
\title{Multimodal feature fusion for CNN-based gait recognition: an empirical comparison}

\author{Francisco M. Castro \and
        Manuel J. Mar\'in-Jim\'enez \and
        Nicol\'as Guil \and
        Nicol\'as P\'erez de la Blanca
}
    
\institute{Francisco M. Castro \at 
              Department of Computer Architecture, University of Malaga, Spain \\
           \and
            Manuel J. Mar\'in-Jim\'enez \at
            Department of Computing and Numerical Analysis, University of Cordoba, Spain \\
            \and
            Nicol\'as Guil \at
            Department of Computer Architecture, University of Malaga, Spain \\
            \and
            Nicol\'as P\'erez de la Blanca \at
            Department of Computer Science and Artificial Intelligence, University of Granada, Spain \\
}
\date{}

\maketitle 
\begin{abstract}
%
People identification in video based on the way they walk (\ie gait) is a relevant task in computer vision using a non-invasive approach. Standard and current approaches typically derive gait signatures from sequences of binary energy maps of subjects extracted from images, but this process introduces a large amount of non-stationary noise, thus, conditioning their efficacy. In contrast, in this paper we focus on the raw pixels, or simple functions derived from them, letting advanced learning techniques to extract relevant features. 
Therefore, we present a comparative study of different Convolutional Neural Network (CNN) architectures by using three different modalities (\ie gray pixels, optical flow channels and depth maps) on two widely-adopted and challenging datasets: TUM-GAID and CASIA-B. In addition, we perform a comparative study between different early and late fusion methods used to combine the information obtained from each kind of modalities.
Our experimental results suggest that 
(\textit{i}) the raw pixel values represent a competitive input modality, compared to the traditional state-of-the-art silhouette-based features (e.g. GEI), since equivalent or better results are obtained; (\textit{ii}) the fusion of the raw pixel information with information from optical flow and depth maps allows to obtain \SOTA results on the gait recognition task with an image resolution several times smaller than the previously reported results; and, (\textit{iii}) the selection and the design of the CNN architecture are critical points that can make a difference between \SOTA results or poor ones. 
\end{abstract}

\section{Introduction} \label{sec:intro}
%
%
The goal of gait-based people identification or simply \textit{gait recognition}, is to identify people by the way they walk. This type of biometric approach is considered non-invasive, since it is performed at a distance, and does not require the cooperation of the subject that has to be identified, in contrast to other methods as iris- or fingerprint-based 
approaches~\cite{ahmadiiris,zeng2018research}. 
Gait recognition has multiple applications in the context of video surveillance, 
ranging from control access in restricted areas to early detection of persons of interest as, for example, v.i.p. customers in a bank office.

From a computer vision point of view, gait recognition could be seen as a particular case of human action recognition~\cite{moeslund2006survey,turaga2008survey}. 
However, gait recognition requires more fine-grained features than action recognition, as differences between different gait styles are usually much more subtle than between common action categories (\eg `high jump' vs. `javelin throw') included in state-of-the-art datasets~\cite{soomro2012ucf101}.

In last years, great effort has been put into the problem of people identification based on gait recognition~\cite{hu2004survey}.
However, previous approaches have mostly used hand-crafted features, as energy maps, after preprocessing video frames by using non-linear filtering. The extracted features, apart from not being easily scalable to diverse datasets, are corrupted by no standard noise derived from the filtering transformation \cite{Han2006pami}. In addition, the noise introduced by the loss of local smoothing between adjacent frames along the temporal-axis makes these features very noisy and variable. Recently, some works based on Convolutional Neural Networks (CNNs) have appeared, for example, Wu~\etal~\cite{Wu2016pami} presents a comparative study of CNN architectures focused on the Gait Energy Image descriptor as input. 

In contrast to all the previous works, we present an approach for gait-based people identification which is independent of any strong image filtering as it uses the raw image, or simple functions derived from it, as input to find the best features (\ie gait descriptor) for the identification task.

The design of our experimental study is directed towards three main objectives. The first objective is the identification of good architectures that, using as input 2D spatial information from a sequence of video frames or 3D spatio-temporal information from a finite subset of video frames, are capable of achieving high scores in the task of gait recognition. To this effect we design 2D-CNN and 3D-CNN architectures with different depth (\ie layers). In addition, as previous works \cite{krizhevsky2012nips} have shown that deeper CNN models achieve better generalisation power than shallower ones, we have also designed a ResNet architecture based on \cite{he2016cvpr}.
%
The second objective is the extraction of the gait signature, which consists of a set of unique characteristics that defines the way of walking of a subject. These characteristics are obtained from different type of input data (\ie appearance, motion and distance), each one defining a different modality (\ie gray-level pixels, optical flow maps and depth maps, respectively).
And, the last objective is to assess if the combination of information derived from different modalities allows to obtain better models for the task of gait recognition. 

To the best of our knowledge, this is the first in-depth study of the impact of CNN architectures and multimodal input data on the gait recognition task using raw input data.

Therefore, the main contributions of this work are:
    \textit{(i)} a comparative study of \SOTA CNN architectures using as input 2D or 3D  information blocks representing spatial and spatio-temporal low-level information, respectively, from data; 
    \textit{(ii)} a thorough experimental study to validate the proposed framework on the standard TUM-GAID and CASIA-B datasets for gait identification; 
    \textit{(iii)} an extensive experimental study of modality fusion; and,
    \textit{(iv)} \SOTA results on both datasets, being our fusion scheme the best approach.

To facilitate the reading of this paper we summarize in Tab.~\ref{tab:abb} a list of abbreviations together with their meaning.

\begin{table}[tbh]
\small
\caption{\textbf{Abbreviations list.} List containing the most used abbreviations along the main text.}
\label{tab:abb}
\begin{center}
\setlength{\tabcolsep}{0.4em} 
\renewcommand{\arraystretch}{1.1} 
\begin{tabular}{|r|c|l|}
\hline
Abbreviation & Topic & Description \\
\hline
CNN & \multirow{9}{*}{General terms} & Convolutional Neural Network \\
OF & & Optical Flow \\
DL & & Deep Learning \\
SVM & & Support Vector Machine \\
MLP & & MultiLayer Perceptron \\
GEI & & Gait Energy Image \\
GAN & & Generative Adversarial Network \\
conv\textit{x} & & Convolutional layer \textit{x} \\
full\textit{x} & & Fully Connected layer \textit{x} \\
\hline
N & \multirow{9}{*}{Dataset scenarios} & Normal scenario in TUM-GAID dataset \\
B & & Bag scenario in TUM-GAID dataset \\
S & & Coating shoes scenario in TUM-GAID dataset \\
TN & & Temporal + Normal scenario in TUM-GAID dataset \\
TB & & Temporal + Bag scenario in TUM-GAID dataset \\
TS & & Temporal + Coating shoes scenario in TUM-GAID dataset \\
nm & & Normal scenario in CASIA-B dataset \\
bg & & Bag scenario in CASIA-B dataset \\
cl & & Coats scenario in CASIA-B dataset \\
\hline
SM-vote & Subsequence combination & SoftMax majority voting strategy \\
SM-prod & strategies & SoftMax product strategy \\
\hline
R1 & \multirow{2}{*}{Metrics} & Rank-1 accuracy metric \\
R5 & & Rank-5 accuracy metric \\
\hline 
\end{tabular}
\end{center}
\end{table}

The rest of the paper is organized as follows. We start by reviewing related work in Sec.~\ref{sec:relwork}. 
Then, Sec.~\ref{sec:approach} explains the different CNN architectures and fusion techniques.
Sec.~\ref{sec:expers} contains the experiments and results. 
Finally, we present the conclusions in Sec.~\ref{sec:conclu}.

\section{Related work}\label{sec:relwork}
%

\subsection{Feature learning}\label{subsec:reldeep}
%

A new realm of this field for recognition tasks started with the advent of Deep Learning (DL) architectures~\cite{bengio2015book}. These architectures are suitable for discovering good features for classification tasks \cite{marin2009wiamis,marin2010icpr,castro2018end} \new{ or system identification in fully connected architectures \cite{de2017stable,de2017usnfis,de2009sofmls,liu2019distributed}}.
Recently, DL approaches based on CNN have been used on image-based tasks with great success \cite{krizhevsky2012nips,simonyan2014corr,zeiler2014eccv}. In the last years, deep architectures for video have appeared, specially focused on action recognition, where the inputs of the CNN are subsequences of stacked frames. The very first approximation of DL applied to stacked frames was proposed in~\cite{le2011cvpr}, where the authors applied a convolutional version of the Independent Subspace Analysis algorithm to sequences of frames. By this way, they obtained low-level features which were used by high-level representation algorithms. A more recent approach was proposed in \cite{karpathy2014cvpr}, where a complete CNN was trained with sequences of stacked frames as input.
In~\cite{simonyan2014nips}, Simonyan and Zisserman proposed to use as input to a CNN a volume obtained as the concatenation of two channels: optical flow in the $x$-axis and $y$-axis. To normalize the size of the inputs, they split the original sequence in subsequences of 10 frames, considering each subsample independently.

Donahue \etal~\cite{donahue2015cvpr} proposed a new viewpoint in DL using a novel architecture called `Long-term Recurrent Convolutional Networks'. This new architecture combined CNN (specialized in spatial learning) with Recurrent Neural Networks (specialized in temporal learning) to obtain a new model able to deal with visual and temporal information at the same time.
Recently, Wang \etal~\cite{wang2015cvpr} combined dense trajectories with DL. The idea was to obtain a powerful model that combined the deep-learnt features with the temporal information of the trajectories. 
They trained a traditional CNN and used dense trajectories to extract the deep features to build a final descriptor that combined the deep information over time. 
On the other hand, Perronnin \etal~\cite{perronnin2015cvpr} proposed a more traditional approach using Fisher Vectors as input to a Deep Neural Network instead of using other classifiers like SVM.
%
Recently, He \etal~\cite{he2016cvpr} proposed a new kind of CNN, named ResNet, which had a large number of convolutional layers and `residual connections' to avoid the vanishing gradient problem.

Although several papers can be found for the task of human action recognition using DL techniques, few works apply DL to the problem of gait recognition. In \cite{hossain2013}, Hossain and Chetty proposed the use of Restricted Boltzmann Machines to extract gait features from binary silhouettes, but a very small probe set (\ie only ten different subjects) were used for validating their approach. 
A more recent work,~\cite{wu2015tmm}, used a random set of binary silhouettes of a sequence to train a CNN that accumulated the calculated features in order to achieve a global representation of the dataset. In~\cite{galai2015cnn}, raw 2D GEI were employed to train an ensemble of CNNs, where a Multilayer Perceptron (MLP) was used as classifier. Similarly, in~\cite{Alotaibi2015aipr} a multilayer CNN was trained with GEI data. A novel approach based on GEI was developed on~\cite{Wu2016pami}, where the CNN was trained with pairs of gallery-probe samples and using a distance metric. Takemura \etal~\cite{takemura2017input} extended this work for the problems of verification and identification using siamese and triplet networks. A different approach was presented in~\cite{he2019multi} where the authors built a multitask generative adversarial network (GAN) for learning view-specific feature representations suitable for the gait recognition problem. Castro \etal~\cite{castro2017iwann} used optical flow obtained from raw data frames. An in-depth evaluation of different CNN architectures based on optical flow maps was presented in~\cite{castro2017biosig}. Finally, in~\cite{marin2017icip} a multitask CNN with a combined loss function with multiple kinds of output labels was presented. 

Recently, some authors have proposed the use of 3D convolutions to extract visual and temporal data from videos. Tran \etal~\cite{tran2015iccv} defined a new network composed of 3D convolutions in the first layers that has been successfully applied to action recognition. Following that idea, Wolf \etal~\cite{Wolf2016icip} built a CNN with 3D convolutions for gait recognition. Due to the high number of parameters that must be trained (3D convolutions implies three times more parameters per convolutional layer), Mansimov \etal~\cite{mansimov2015arxiv} showed several ways to initialize a 3D CNN from a 2D CNN.

Despite most CNNs are trained with visual data (e.g. images or videos), there are some works that build CNNs for different kinds of data like inertial sensors or human skeletons. Holden \etal~\cite{holden2015} proposed a CNN that corrected wrong human skeletons obtained by other methods or devices (e.g. Microsoft Kinect). Neverova \etal~\cite{neverova2015arxiv} built a temporal network for active biometric authentication with data provided by smartphone sensors (e.g. accelerometers, gyroscope, etc.). Delgado-Esca\~no \etal~\cite{delgado2019end} built an end-to-end CNN that used several inertial sensors to produce multiple biometric outputs such as subject id, gender or age.
%
%

\subsection{Information fusion}\label{subsec:relfusion}
Since there are different ways or modalities for representing the same data, an interesting idea would be to try to combine those modalities into a single one that could benefit from the original information. To perform this task, several methods have appeared~\cite{atrey2010ms,wu2009es}. Also, the emergence of new cheaper devices that record multimodal spectrums (e.g. RGB, depth, infrared) has allowed to investigate how to fuse that information to build richer and more robust representations for the gait recognition problem. 
Traditionally, fusion methods are divided into \textit{early fusion} methods (or feature fusion) and \textit{late fusion} (or decision fusion). The first ones try to build descriptors by fusing descriptors of different modalities, frequently, using the concatenation of the descriptors into a bigger one as in~\cite{chai2015paa}. On the other hand, late fusion tries to fuse the decisions obtained by each classifier of each modality, usually, by applying arithmetic operations like sums or products on the scores obtained by each classifier as in~\cite{chai2015paa, hofmann2014tumgaid}. Castro \etal~\cite{castro2015caip,castro2016mva} perform an extensive comparative between late fusion and early fusion methods including the traditional fusion schemes and others more grounded that can perform robust fusions.
Fusion has also been employed with CNNs to improve the recognition accuracy for different computer vision tasks. For example, two independent CNNs fed with optical flow maps and appearance information (\ie RGB pixel volumes) were employed in~\cite{simonyan2014nips} to perform action recognition. Then, class score fusion is used to combine the softmax output of both CNNs. In a similar way, Eitel \etal~\cite{Eitel2015} proposed a DL approach for object recognition by fusing RGB and depth input data. They concatenated the outputs of the last fully-connected layers of both networks (those processing RGB and depth data) and processed them through an additional fusion layer. Wang \etal~\cite{WangAnran2015} also employed a multimodal architecture composed by two CNN networks to process RGB-D data. They proposed to learn two independent transformations of the activations of the second fully-connected layer of each network, so correlation of color and depth features was maximized. In addition, these transformations were able to improve the separation between samples belonging to different classes.    

In this work, we explore several fusion techniques for the problem of gait-based people identification, combining automatically-learnt gait signatures extracted from gray pixels, optical flow and depth maps.

%

\section{Proposed approach} \label{sec:approach}
%
\begin{figure*}[t]
\begin{center}
   \includegraphics[width=0.98\textwidth]{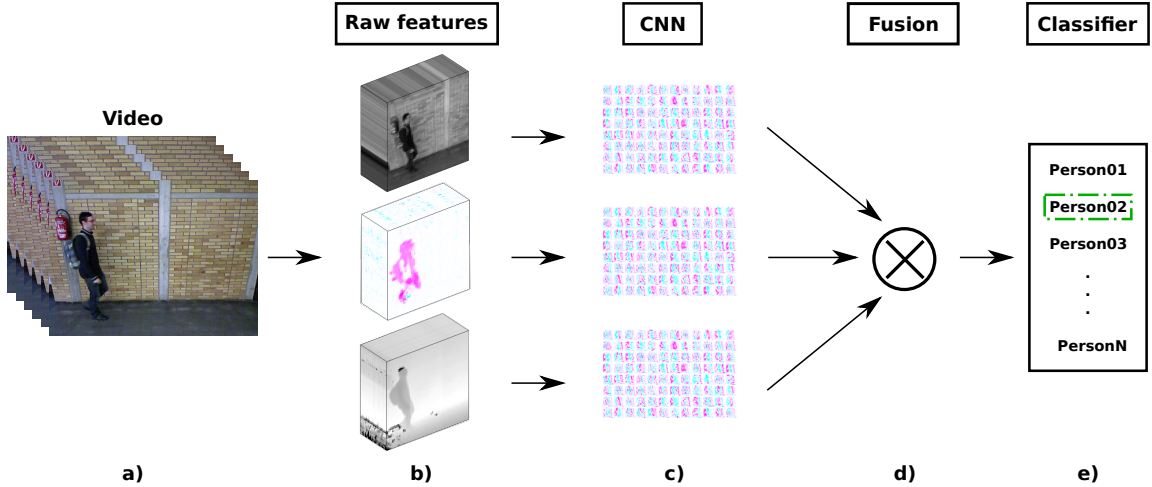}
\end{center}
\reduceFigSpace
\caption{\textbf{Pipeline for gait recognition}. A scheme of the proposed approach assuming a video sequence composed by RGB-D frames is being processed. 
In a) the input video sequence is shown. In b) three modalities are extracted (gray, optical flow and depth) from a subsequence of the video and stacked volumes for each modality are built. In c) gait signatures calculated by the CNN after processing stacked volumes are depicted. Then, CNN fusion is carried out in d). In e) identity of the subject appearing in the video is calculated taking into account all the subsequences processed from the input video.}
\label{fig:pipeline}
\end{figure*}

In this section we describe our proposed framework to address the problem of gait recognition using CNNs. 
The pipeline proposed for gait recognition based on CNNs is represented in Fig.~\ref{fig:pipeline}: 
\textit{(a)} extract consecutive frames from the video sequence;
 \textit{(b)} gathering different modalities along the whole sequence and building up a data cuboid from consecutive modality maps; 
 \textit{(c)} feeding the CNN with the modality cuboid to extract the gait signature; 
 \textit{(d)} fusing information from the different modalities; and, 
 \textit{(e)} applying a classifier to decide the subject identity. 


The datasets employed in this work provide both RGB and RGB-D video frames. From RGB components we have generated two modalities: {\it gray pixels} and {\it optical flow}. If distance information is available (as in RGB-D cameras) we also employ an additional modality called {\it depth}.

\subsection{Input data} \label{subsec:features}
\begin{figure}[t]
\begin{center}
   \includegraphics[width=0.49\linewidth]{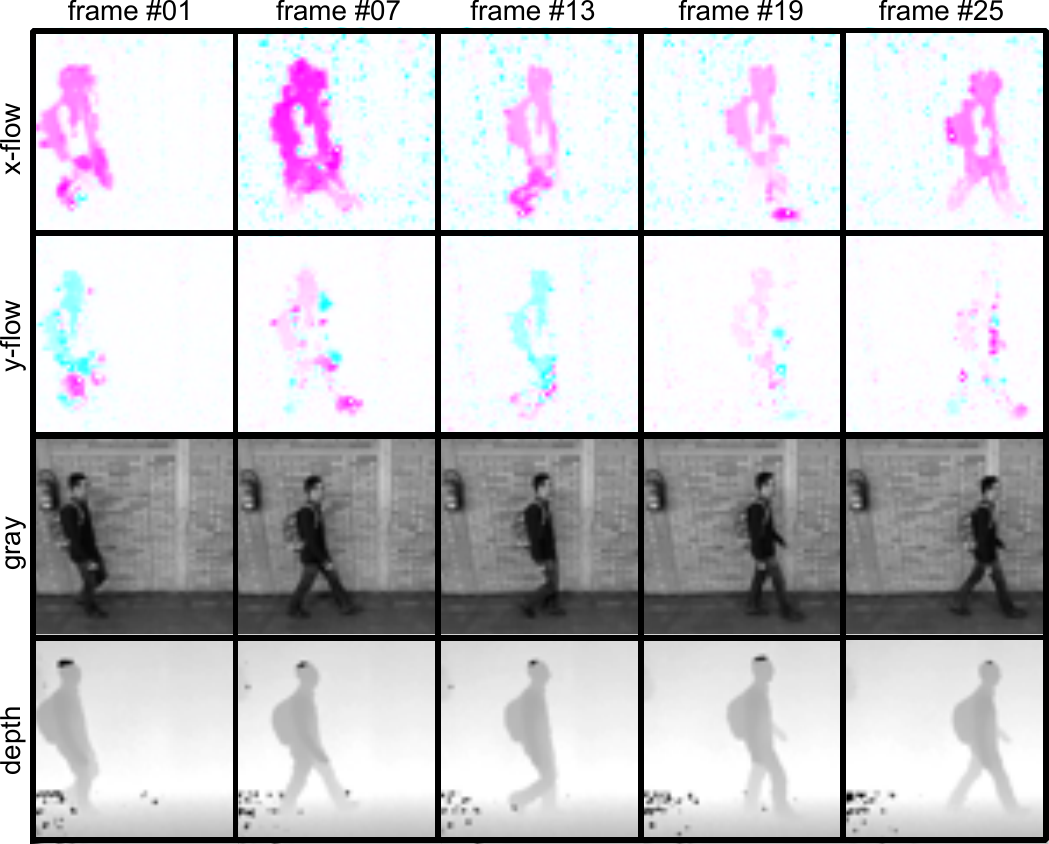}
\end{center}
\caption{\textbf{CNN input data}. Sample frames extracted from a subsequence of $25$ frames. \textbf{(top rows)} Optical flow in $x$-axis and $y$-axis. where positive flows are displayed in pink and negative flows in blue (best viewed in color). 
\textbf{(bottom rows)} Gray pixels and depth maps of the same sequence.}
\label{fig:subseqs}
\end{figure}

We describe here the different types of modalities used as input for the proposed CNN architecture. 
In particular, we use optical flow, gray pixels and depth maps since they provide different type of information, are easy to compute and are available in the used datasets.
Our intuition is that optical flow will focus on describing the \emph{gait-related motion} as shown in~\cite{castro2017iwann}, gray pixels are widely used in deep learning~\cite{krizhevsky2012nips} to encode \emph{appearance} information and finally depth maps will provide some soft geometric information as commented in~\cite{castro2016mva}.
An example of the three types of modalities is represented in Fig.~\ref{fig:subseqs}.


\subsubsection{Optical flow} \label{subsub:OF}
The use of optical flow (OF) as input data for action representation in video with CNN has already shown excellent results~\cite{simonyan2014nips}.
Nevertheless human action is represented by a wide, and usually well defined, set of local motions. In our case, the set of motions differentiating one gait style from another is much more subtle and local. 

Let $F_t$ be an OF map computed at time $t$ and, therefore, $F_t(x,y,c)$ be the value of the OF vector component $c$ located at coordinates $(x,y)$, where $c$ can be either the horizontal or vertical component of the corresponding OF vector. 
The input data $I_L$ for the CNN are cuboids built by stacking $L$ consecutive OF maps $F_t$, where $I_L(x,y,2k-1)$ and $I_L(x,y,2k)$ corresponds to the value of the horizontal and vertical OF components located at spatial position $(x,y)$ and time $k$, respectively, ranging $k$ in the interval $[1,L]$.

Since each original video sequence will probably have a different temporal length, and a CNN requires a fixed size input, we extract subsequences of $L$ frames from the full-length sequences. 
In Fig.~\ref{fig:subseqs} we show five frames distributed every six frames along a subsequence of twenty-five frames in total (\ie frames 1, 7, 13, 19, 25).
The first row shows the horizontal component of the OF ($x$-axis displacement) and second row shows the vertical component of the OF ($y$-axis displacement). It can be observed that most of the motion flow is concentrated in the horizontal component, due to the displacement of the person.
In order to remove noisy OF located in the background, as it can be  observed in Fig.~\ref{fig:subseqs}, we might think in applying a preprocessing step for filtering out those vectors whose magnitude is out of a given interval. However, since our goal in this work is to minimize the manual intervention in the process of gait signature extraction, we will use those OF maps as returned by the OF algorithm.


\subsubsection{Gray-level pixels} \label{subsub:pixels}
When using CNNs for object detection and categorization, the most popular modality is raw pixels~\cite{krizhevsky2012nips}. 
In contrast to \cite{simonyan2014nips}, that uses single RGB frames for action recognition, we build cuboids of gray pixels with the aim of better capturing the important features of the subject appearance. 
Note that in gait recognition, color is not as informative as it is for object recognition. Therefore, using only gray intensity will eventually help CNN to focus just on the gait-relevant information. An example can be seen in the corresponding row of Fig.~\ref{fig:subseqs}.

\subsubsection{Depth maps} \label{subsub:depth}
As far as we know, the use of depth information has not been explored much in the field of gait recognition. In \cite{hofmann2014tumgaid} they basically used depth to segment people from background and compute the \textit{Gait Energy Volume} descriptor~\cite{sivapalan2011gev}. Castro \etal~\cite{castro2016mva} represented depth information in a gray-scale image where the intensity of a pixel is the depth value scaled to $[0, 255]$. 
In our opinion, depth information is rich and should be studied in depth for this problem.
Therefore, given a sequence of depth maps, we extract depth volumes that will be used as input data for the corresponding CNN architecture.
An example of depth maps can be seen in the bottom row of Fig.~\ref{fig:subseqs}.

\subsection{CNN architectures for single modality} \label{subsec:cnnarch}
\begin{figure}[t]
\begin{center}
   \includegraphics[width=0.99\linewidth]{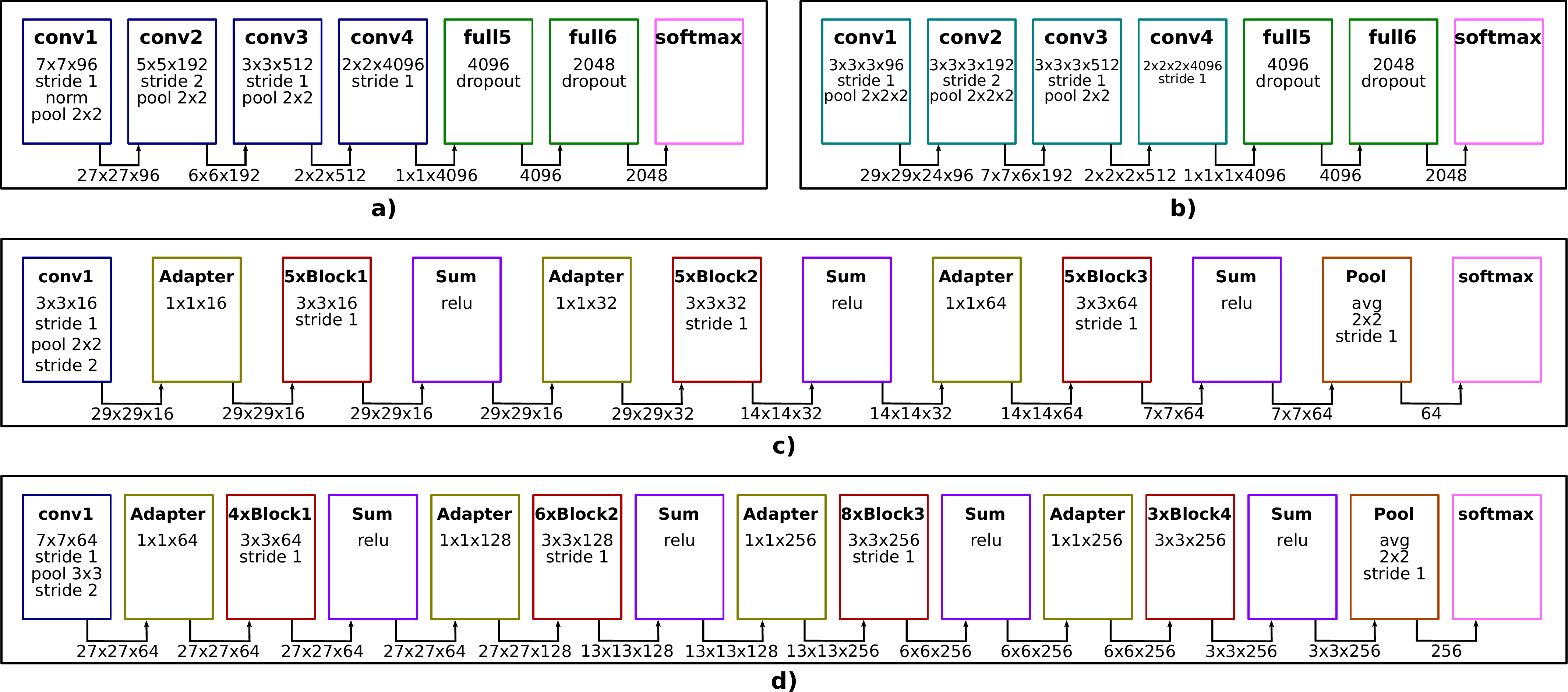}
\end{center}
\caption{\textbf{Proposed CNN architectures for gait signature extraction}.\textbf{a) 2D-CNN:} linear CNN with four 2D convolutions, two fully connected layers and a softmax classifier. \textbf{b) 3D-CNN:} 3D CNN four 3D convolutions, two fully connected layers and a softmax classifier. \textbf{c) ResNet-A:} residual CNN with a 2D convolution, three residual blocks (red boxes), an average pooling layer and a final softmax classifier. \textbf{d) ResNet-B:} residual CNN with a 2D convolution, four residual blocks (red boxes), an average pooling layer and a final softmax classifier. Arrows connecting two consecutive boxes show the tensor shape \new{$Height\times Width\times Channels$} (output for the left box and input for the right one). \new{All initial inputs are $60\times 60 \times 50$}. More details in the main text.
}
\label{fig:archCNN}
\end{figure}



We have selected the three architectures that most frequently appear in the bibliography and produce state-of-the-art results in different topics (e.g. action recognition, object detection, etc.). The three proposed architectures are: 
\textit{(i)}  a linear CNN with 2D convolutions (\textit{2D-CNN}) based on AlexNet~\cite{krizhevsky2012nips}, which is the traditional and most common architecture; 
\textit{(ii)}  a linear CNN with 3D convolutions and pooling (\textit{3D-CNN}), which is specially designed to capture information in videos~\cite{ji20123d}; and, \textit{(iii)} a 2D very deep residual CNN (\textit{ResNet}~\cite{he2016cvpr}), which produces state-of-the-art results in most challenging tasks. 

The input to our CNN is a volume of gray pixels, depth maps or OF channels with size $N \times N \times L$ for gray and depth, and size $N \times N \times 2L$ for OF since it has two components ($x$ and $y$). We refer the reader to Sec.~\ref{subsec:impldets} for the actual values of $N$ and $L$ used in the experiments.
Note that for the case of the 3D CNN on OF, the input must be split into two blocks of size $N\times N \times L$ to apply the temporal filters at each OF channel independently, which will be concatenated at deeper layers.

We describe below the four particular models compared in the experimental section (Sec.~\ref{sec:expers}). Note that we use the term `softmax layer' to refer to a fully-connected layer with as many units as classes followed by a softmax exponential layer. Moreover, in all models, the \textit{gait signatures} are extracted  from the layer preceding this softmax layer.


\new{The most common layers of these architectures are the convolutions, pooling and ReLU activation function described in Eq.~\ref{eq:layers}.}

\new{\begin{equation}
    \begin{aligned}
    \mathrm{conv}(\mathbf{x}, \mathbf{w}, \mathbf{b}) = & \sum_i w_i \cdot x_i + \mathbf{b} \\
        \mathrm{ReLU}(x_i) = & \mathrm{max}(0, x_i) \\
        \mathrm{pooling}(\mathbf{x_\omega}) = & \mathrm{max}(\mathbf{x_\omega}) 
    \end{aligned}
    \label{eq:layers}
\end{equation}}

\new{where $\mathbf{x}$ is an input matrix, $x_i$ is an element of that matrix, $w_i$ is a weight of the convolutional filter, $\mathbf{b}$ is the bias term, $\mathrm{max}(\cdot)$ is the maximum function and $\mathbf{x_\omega}$ represents the region of $\mathbf{x}$ where the pooling operation is applied.}

\myparagraph{2D-CNN:}
This CNN is composed of the following sequence of layers (Fig.~\ref{fig:archCNN}a): 
`\textit{conv1}', 96 filters of size $7\times 7$ applied with stride 1 followed by a normalization and max pooling $2\times 2$;
`\textit{conv2}', 192 filters of size $5\times 5$ applied with stride 2 followed by max pooling $2\times 2$;
`\textit{conv3}', 512 filters of size $3\times 3$ applied with stride 1 followed by max pooling $2\times2$;
`\textit{conv4}', 4096 filters of size $2\times 2$ applied with stride 1;
`\textit{full5}', fully-connected layer with 4096 units and dropout;
`\textit{full6}', fully-connected layer with 2048 units and dropout; and,
`\textit{softmax}', softmax layer with as many units as subject identities.
All convolutional layers use the rectification (ReLU) activation function.

\myparagraph{3D-CNN:}
As optical flow has two components and the CNN uses temporal kernels, the network is split into two branches: $x$-flow and $y$-flow. Therefore, each branch contains half of the total filters described below. Then, this CNN is composed by the following sequence of layers (Fig.~\ref{fig:archCNN}b): 
`\textit{conv1}', 96 filters of size $3\times 3 \times 3$ applied with stride 1 followed by a max pooling $2\times 2 \times 2$;
`\textit{conv2}', 192 filters of size $3\times 3 \times 3$ applied with stride 2 followed by max pooling $2\times 2 \times 2$;
`\textit{conv3}', 512 filters of size $3\times 3 \times 3$ applied with stride 1 followed by max pooling $2\times2 \times 2$;
`\textit{conv4}', 4096 filters of size $2\times 2 \times 2$ applied with stride 1;
`\textit{concat}', concatenation of both branches ($x$-flow and $y$-flow);
`\textit{full5}', fully-connected layer with 4096 units and dropout;
`\textit{full6}', fully-connected layer with 2048 units and dropout; and,
`\textit{softmax}', softmax layer with as many units as subject identities.
All convolutional layers use the rectification (ReLU) activation function. %

\myparagraph{ResNet-A: }
This CNN is composed by the following sequence of layers and residual blocks (a sequences of two convolutions of size $3 \times 3$, as defined in~\cite{he2016cvpr} for CIFAR Dataset). This model is specially designed for small datasets with low variability, as this kind of networks tends to overfit due to its high number of layers. As our architecture follows the indications defined by the authors~\cite{he2016cvpr}, we only describe the main blocks (Fig.~\ref{fig:archCNN}c): 
`\textit{conv1}', 16 filters of size $3\times 3$ applied with stride 1 followed by a  max pooling $2\times 2$ and stride 2;
`\textit{block 1}', 5 residual blocks with convolutions of 16 filters of size $3\times 3$ applied with stride 1;
`\textit{block 2}', 5 residual blocks with convolutions of 32 filters of size $3\times 3$ applied with stride 1;
`\textit{block 3}', 5 residual blocks with convolutions of 64 filters of size $3\times 3$ applied with stride 1;
`\textit{average pooling}', size $8\times 8$ with stride 1; and,
`\textit{softmax}', softmax layer with as many units as subject identities. 
All convolutional layers use the rectification (ReLU) activation function and batch normalization.

\myparagraph{ResNet-B: }
This model is an extension of the previous model ResNet-A. The number and size of layers of this model is increased and is specially designed for datasets with high variability (e.g. CASIA-B). This CNN is composed by the following sequence of layers and residual blocks (a sequence of three convolutions of size $1 \times 1$, $3 \times 3$ and $1 \times 1$, as defined in~\cite{he2016cvpr}). As our architecture follows the indications defined by the authors, we only describe the main blocks (Fig.~\ref{fig:archCNN}d): 
`\textit{conv1}', 64 filters of size $7\times 7$ applied with stride 1 followed by a  max pooling $3\times 3$ and stride 2;
`\textit{block 1}', 4 residual blocks with convolutions of 64 filters of size $3\times 3$ applied with stride 1;
`\textit{block 2}', 6 residual blocks with convolutions of 128 filters of size $3\times 3$ applied with stride 1;
`\textit{block 3}', 8 residual blocks with convolutions of 256 filters of size $3\times 3$ applied with stride 1;
`\textit{block 4}', 3 residual blocks with convolutions of 256 filters of size $3\times 3$ applied with stride 1;
`\textit{average pooling}', size $2\times 2$ with stride 1; and,
`\textit{softmax}', softmax layer with as many units as subject identities.
All convolutional layers use the rectification (ReLU) activation function and batch normalization.

\subsubsection{Model training}\label{subsec:implemCNN}

For 2D and 3D models, we perform a curriculum learning strategy~\cite{bengio2009curriculum} to speed up and to facilitate the convergence. In this learning process, initially, we train a simplified version of each model (\ie less units per layer and no dropout) and, then, we use the learned network parameters for initializing the layers of a more complex architecture (\ie 0.1 dropout and more filters and units). This learning process is applied three times until we develop the fourth model. This final model is shown in Fig.~\ref{fig:archCNN} and described in the previous section.
%

%
During CNN training, the network parameters are learnt using the mini-batch stochastic gradient descent \new{(SGD) algorithm, described in Eq.~\ref{eq:sgd}.}

\new{\begin{equation}
\begin{aligned} 
    \Delta\theta_t = & \gamma \cdot \Delta\theta_{t-1} + \alpha \frac{1}{n}\sum^{n}_{i=1}(h_\theta(x^{(i)})-y^{(i)}) \cdot x^{(i)} \\
    \theta \leftarrow & \theta - \Delta\theta_t
\end{aligned}
\label{eq:sgd}
\end{equation}}

\new{where $\theta$ are the trainable parameters of the model, $\alpha$ is the learning rate, $n$ is the size of the mini-batch, $h_\theta(x^{(i)})$ is the output value of the model for the $x^{(i)}$ sample, $y^{(i)}$ is the label of the sample $i$, $x^{(i)}$ is the input sample, $\gamma$ is the momentum constant, $\Delta\theta_t$ is the current weights update and $\Delta\theta_{t-1}$ is the previous weights update.}

\new{In our case, we set momentum} equal to $0.9$ in the first two curriculum learning iterations of the 2D and 3D models, and 0.95 during the last one. Note that ResNet-A and ResNet-B are trained from scratch in just one stage (without curriculum learning strategy) so momentum for these networks is set to $0.9$. We set weight decay to $5 \cdot 10^{-4}$ and dropout to $0.4$ (when corresponds). The number of epochs is limited to 20 in TUM-GAID, and the learning rate is initially set to $10^{-2}$, which is divided by ten when the validation error plateaus. Due to the nature of the ResNet models, the initial learning rate is set to $0.1$.

In CASIA-B we limit the training stage to 30 epochs, the learning rate is initially set to $10^{-3}$ and it is divided by two when the validation error gets stuck.
At each epoch, a mini-batch of 150 samples %
is randomly selected from a balanced training set (\ie almost the same proportion of samples per class). Note that for ResNet models we use a mini-batch of 64 samples.
When the CNN training has converged, we perform five more epochs on the joint set of training and validation samples.
%

To run our experiments we use the implementation of CNN provided in MatConvNet library~\cite{vedaldi2015matconvnet}. This library allows to develop CNN architectures in an easy and fast manner using the Matlab environment. In addition, it takes advantage of CUDA and cuDNN~\cite{chetlur2014cudnn} to improve the performance of the algorithms.
Using this open source library will allow other researchers to use our trained models and reproduce our experimental results.

\subsubsection{Gait signature evaluation}\label{subsec:singleclasif}
%
Once the model is trained, we can extract the gait signatures for a given input. The final stage consists on classifying those signatures to derive a subject identity. \new{Note that, in order to obtain the probabilities from the gait signatures, we use the well-known softmax layer described in Eq.~\ref{eq:softmax}}.

\new{\begin{equation}
    \mathrm{softmax}(x_i) = \frac{e^{x_i}}{\sum_j e^{x_j}}
    \label{eq:softmax}
\end{equation}}

\new{where $x_i$ is an element of the gait signature.}

As it was explained in Sec.~\ref{subsec:features}, we split the whole video sequence into overlapping subsequences of a fixed length, and those subsequences are classified independently. Then, we combine the identification results obtained from each subsequence to produce the identification result of the whole video sequence. In our experiments we implement two different strategies for combining subsequence results:
\begin{description}
\item{SM-Vote}. After softmax decision we apply a \textit{majority voting} strategy on the labels assigned to each subsequence.
\item{SM-Prod}. The identity is derived from the product of softmax vectors (\ie probability distributions $P_i$) obtained:
\begin{equation}\label{eq:prodsoftmax}
   P(v=c) = \prod_{i=1}^t P_i(s_i = c) ,   
\end{equation}%
where $t$ is the number of subsequences extracted from video $v$, $P(v=c)$ is the probability of assigning the identity $c$ to the person in video $v$ and $P_i(s_i = c)$ is the probability of assigning the identity $c$ to the person in subsequence $s_i$. 
\end{description}



\subsection{Multiple modalities} \label{subsec:multiclasif}
In this case, we explore different fusion techniques for the signatures extracted from different modalities, expecting that fused information improves the subject identification task.

\myparagraph{Late fusion.}
Focusing on the softmax scores returned by each CNN,
we explore the following approaches to combine them: product and weighted sum.
These approaches are considered as `late fusion' ones, as fusion is performed on the classification scores.

%

\noindent\textit{A) Product of softmax vectors.} %
Given a set of $n$ softmax vectors $\{P_i\}$ obtained from a set of different modalities $\{m_i\}$, 
a new score vector $S_{\mathrm{prod}}$ is obtained as:
\begin{equation}\label{eq:multimodalProd}
   S_{\mathrm{prod}}(v=c) = \prod_{i=1}^n P_i(m_i = c)    
\end{equation}%
where $n$ is the number of modalities used to classify video $v$, 
$ S_{\mathrm{prod}}(v=c)$ can be viewed as the probability of assigning the identity $c$ to the person in video $v$ and $P_i(m_i = c)$ is the probability of assigning the identity $c$ to the person in modality $m_i$.

\noindent\textit{B) Weighted sum of softmax vectors.} %
Given a set of $n$ softmax vectors obtained from a set of different modalities $\{m_i\}$ a new score vector $S_{\mathrm{ws}}$ is obtained as:
\begin{equation}\label{eq:multimodalWS}
   S_{\mathrm{ws}}(v=c) = \sum_{i=1}^n \beta_i  P_i(m_i = c) ,   
\end{equation}%

where $n$ is the number of modalities used to classify video $v$, 
$ S_{\mathrm{ws}}(v=c)$ can be viewed as the probability of assigning the identity $c$ to the person in video $v$, $P_i(m_i = c)$ is the probability of assigning the identity $c$ to the person in modality $m_i$ and $\beta_i$ is the weight associated to modality $m_i$, subject to $\beta_i > 0$ and $\sum_{i=1}^n \beta_i = 1$.

$\beta$ values are selected empirically by cross-validation. By this way, we have split the training set into training and validation subsets to try different combinations of $\beta$ values. In our case, we have tested all possible values in the range $[0.1, 0.9]$ with steps of $0.1$. Note that the values used for each experiment are specified in its corresponding section. Moreover, the same process is followed during training and test time since we are using late fusion techniques that do not affect the training process.

\begin{figure}[t]
\begin{center}
   \includegraphics[width=0.58\linewidth]{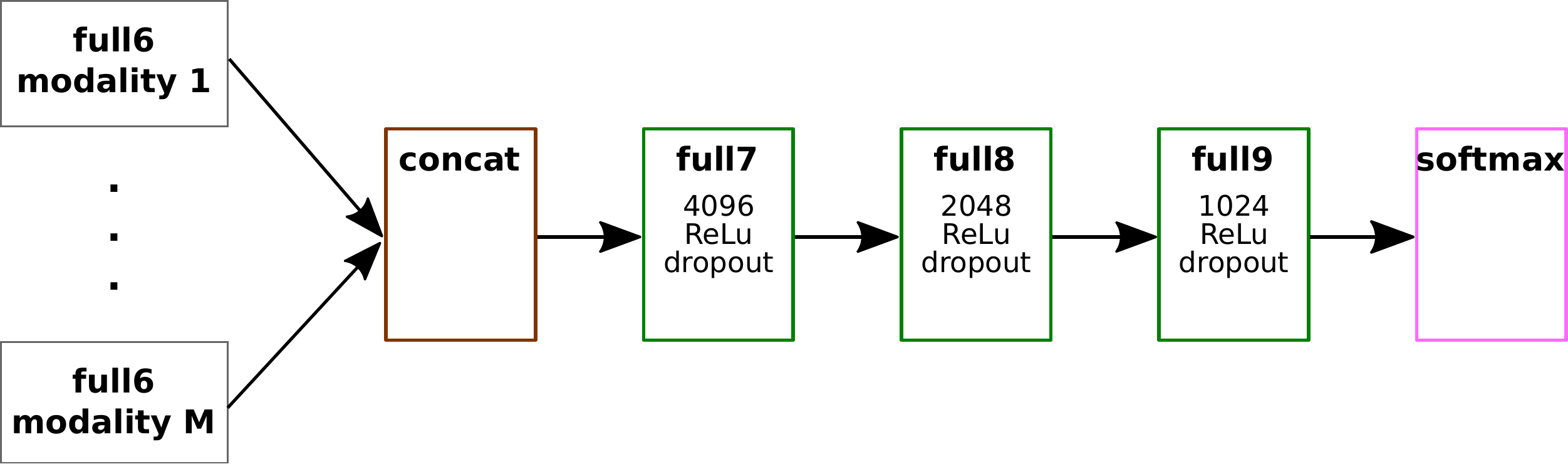}
\end{center}
\caption{\textbf{Proposed set of layers for early fusion}. A concatenation layer and three fully-connected layers are followed by a softmax classifier used to directly derive an identity. More details in the main text.}
\label{fig:archNN}
\end{figure}

\vspace{0.5cm}
\myparagraph{Early fusion.}
The fusion performed at gait signature level is known as `early fusion'. 
In our case, as we are working with CNNs, early fusion could be performed at any layer before the `softmax' one. Depending on the layer, the fused descriptors are matrices (fusion before a convolutional layer) or vectors (fusion before a fully-connected layer). We have tried all the possible fusion locations for our CNNs and we have selected the best solution according to the results obtained. In our case, the best early fusion location is after layer `full6' of each modality. The activations of those layers are concatenated and fed into a new set of layers to perform the actual fusion. Therefore, we extend the 2D and 3D networks shown in Fig.~\ref{fig:archCNN} with the set of additional layers summarized in Fig.~\ref{fig:archNN}: 
\begin{description}
\setlength\itemsep{-0.1cm} 
    \item[`\textit{concat}:'] concatenation layer;
    \item[`\textit{full7}:'] fully-connected layer with 4096 units, ReLU and dropout;
    \item[`\textit{full8}:'] fully-connected layer with 2048 units, ReLU and dropout;
    \item[`\textit{full9}:'] fully-connected layer with 1024 units, ReLU and dropout; and,
    \item[`\textit{softmax}:'] softmax layer with as many units as subject identities.
\end{description}
During the training process, the parameters of the whole CNN (\ie the branch of each modality and the fusion layers) are trained altogether, automatically learning the best combination of network parameters for the modalities. At test time, we use the learnt network parameters to fuse the information from the new samples.
From our point of view, this kind of fusion is considered early as it is not performed at classification-score level, as done above.

For ResNet models, due to their high number of layers, we do not stack more fully-connected layers to prevent overfitting. Therefore, the selected early fusion architecture is the same as for the rest of models but without fully-connected layers:
\begin{description}
\setlength\itemsep{-0.1cm} 
    \item[`\textit{concat}:'] concatenation layer;
    \item[`\textit{softmax}:'] softmax layer with as many units as subject identities and dropout.
\end{description}

\section{Experiments and results} \label{sec:expers}
%

We present here the experiments designed to validate our approach and the results obtained on the selected datasets for gait recognition.

\subsection{Datasets} \label{subsec:datasets}
We run our experiments on two widely used and challenging datasets for gait recognition: TUM-GAID~\cite{hofmann2014tumgaid} and  CASIA-B~\cite{yu2006casia}.
Both datasets are described below.

%

\begin{figure}[tbh!]
\begin{center}
   \includegraphics[width=0.98\linewidth]{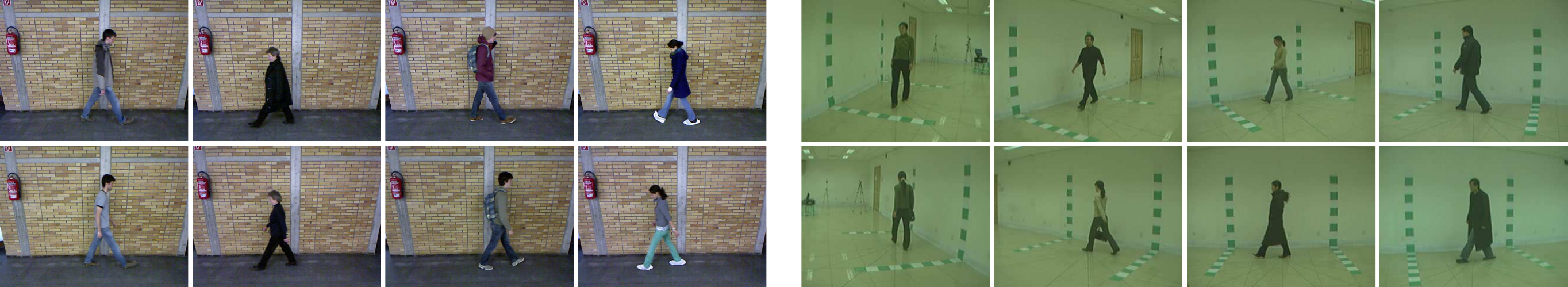}
\end{center}
\caption{\textbf{Datasets for gait recognition}. \textbf{(left) TUM-GAID.} People walking indoors under four walking conditions: normal walking, wearing coats, carrying a bag and wearing coating shoes. Top and bottom rows show the same set of subjects but in different months of the same year.
\textbf{(right) CASIA-B.} People walking indoors recorded from eleven camera viewpoints and under three walking conditions: normal walking, wearing coats and carrying a bag.
}
\label{fig:datasets}
\end{figure}

\myparagraph{TUM-GAID.}
In `TUM Gait from Audio, Image and Depth' (TUM-GAID) 305 subjects perform two walking trajectories in an indoor environment. %
The first trajectory is performed from left to right and the second one from right to left. Therefore, both sides of the subjects are recorded. Two recording sessions were performed, one in January, where subjects wore heavy jackets and mostly winter boots, and the second in April, where subjects wore different clothes. 
The action is captured by a Microsoft Kinect sensor which provides a video stream with a resolution of $640 \times 480$ pixels with a frame rate of approximately 30 fps.
Some examples can be seen in the left part of Fig.~\ref{fig:datasets} depicting the different conditions included in the dataset.

Hereinafter the following nomenclature is used to refer each of the four walking conditions considered: \textit{normal} walk (\textit{N}), carrying a \textit{backpack} of approximately 5 kg (\textit{B}), wearing coating \textit{shoes} (\textit{S}), as used in clean rooms for hygiene conditions, 
and \textit{elapsed time} (\textit{TN-TB-TS}). 
Each subject of the dataset is composed of: six sequences of normal walking (\textit{N1, N2, N3, N4, N5, N6}), two sequences carrying a bag (\textit{B1, B2}) and two sequences wearing coating shoes (\textit{S1, S2}). In addition, 32 subjects were recorded in both sessions  (\ie January and April) so they have 10 additional sequences (\textit{TN1, TN2, TN3, TN4, TN5, TN6, TB1, TB2, TS1, TS2}). Therefore, the overall amount of videos is 3400.

We follow the experimental protocol defined by the authors of the dataset \cite{hofmann2014tumgaid}. Three subsets of subjects are proposed: training, validation and testing. The training set is used for obtaining a robust model against the different covariates of the dataset. This partition is composed of 100 subjects and the sequences \textit{N1} to \textit{N6}, \textit{B1}, \textit{B2}, \textit{S1} and \textit{S2}. The validation set is used for validation purposes and contains 50 different subjects with the sequences \textit{N1} to \textit{N6}, \textit{B1}, \textit{B2}, \textit{S1} and \textit{S2}. Finally, the test set contains other 155 different subjects used in the test phase. As the set of subjects is different between the test set and the training set, a new training of the identification model must be performed. 
For this purpose, the authors reserve the sequences \textit{N1} to \textit{N4}, from the subject test set, to train the model again and the rest of sequences are used for testing and to obtain the accuracy of the model. In the \textit{elapsed time} experiment, the temporal sequences (\textit{TN1, TN2, TN3, TN4, TN5, TN6, TB1, TB2, TS1, TS2}) are used instead of the normal ones and the subsets are: 10 subjects in the training set, 6 subjects in the validation set and 16 subjects in the test set.
%

In our experiments, after parameter selection, the validation sequences are added to the training set for fine-tuning the final model.

\myparagraph{CASIA-B.}
In CASIA-B 124 subjects perform walking trajectories in an indoor environment (right part of Fig.~\ref{fig:datasets}). The action is captured from 11 viewpoints (\ie from $0^o$ to $180^o$ in steps of $18^o$) with a video resolution of $320 \times 240$ pixels. Three situations are considered: normal walk (\textit{nm}), wearing a coat (\textit{cl}), and carrying a bag (\textit{bg}).
%
The authors of the dataset indicate that sequences 1 to 4 of the `\textit{nm}' scenario should be used for training the models. Whereas the remaining sequences should be used for testing: sequences 5 and 6 of `\textit{nm}', 1 and 2 of `\textit{cl}' and 1 and 2 of `\textit{bg}'. Therefore, we follow this protocol in our experiments, unless otherwise stated.
This makes a total of 496 video sequences for training, per camera viewpoint.

\subsection{Implementation details}\label{subsec:impldets}
%
We ran our experiments on a computer with 32 cores at 2.3 GHz, 256 GB of RAM and a GPU NVidia Titan X Pascal, 
with MatConvNet library~\cite{vedaldi2015matconvnet} running on Matlab 2016a for Ubuntu 16.04. 

For the following experiments with CNN, we resized all the videos to a common resolution of $80 \times 60$ pixels, keeping the original aspect ratio of the video frames.
Preliminary experiments support this choice~\cite{castro2017iwann}, as this size exhibits a good trade-off between computational cost and recognition performance.
Note that the resolution $80 \times 60$ is 4 times lower than the original CASIA-B one and 8 times lower than the TUM-GAID one.

Given the resized video sequences, we compute dense \textit{OF} on pairs of frames by using the method of Farneback \cite{Farneback03} implemented in OpenCV library~\cite{opencv_library}. In parallel, people are located in a rough manner along the video sequences by background subtraction~\cite{kaewtrakulpong2002bmm}. 
Then, we crop the video frames to remove part of the background with $N=60$, obtaining video frames of $60\times 60$ pixels (full height is kept) and to align the subsequences (people are $x$-located in the middle of the central frame, \#13) as in Fig.~\ref{fig:subseqs}.

Finally, from the cropped \textit{OF} maps, we build subsequences of $L=25$ frames by stacking \textit{OF} maps with an overlap of $\mathcal{O}\%$ frames. In our case, we chose $\mathcal{O}=80\%$, that is, to build a new subsequence, we use $20$ frames of the previous subsequence and $5$ new frames. For most state-of-the-start datasets, 25 frames cover almost one complete gait cycle, as stated by other authors~\cite{barnich2009prl}. 
Therefore, each \textit{OF} volume has size $60\times 60 \times 50$.

The same process described above is applied to the gray pixels and depth inputs with values $N=60$ and $L=25$, obtaining  volumes of size $60\times 60 \times 25$. Before feeding the CNN with those data volumes, the mean of the training set for each modality is subtracted to the input data.
Both gray and depth values are normalized to the range $[0,255]$. Note that in CASIA-B, due to the high variability between viewpoints, it is necessary to normalize gray values to the range $[0, 1]$.


To increase the amount of training samples we add mirror sequences and apply spatial displacements of $\pm 5$ pixels in each axis, obtaining a total of 8 new samples from each original sample.

\subsection{Performance evaluation} \label{subsec:metrics}
For each test sequence, we return a sorted list of possible identities, where the top one identity corresponds to the largest scored one. According to this we use two quantitative metrics, \textit{rank-1} (R1) and \textit{rank-5} (R5) accuracy, to measure the performance of the proposed system. Specifically, metric \textit{rank-1} measures the percentage of test samples where the top one assigned identity corresponds to the correct one. Metric \textit{rank-5} measures the percentage of test samples where the ground truth identity is included in the first five ranked identities for the corresponding test sample. Note that \textit{rank-5} is less strict than \textit{rank-1} and, in a real system, it would allow to verify if the target subject is any of the top five most probably ones.
In order to compute the accuracy metric, we use the general equation:
\begin{equation}
    accuracy = \frac{\#correct\_predictions}{\#total\_predictions} \cdot 100,
\end{equation}
where the numerator indicates the number of correct predictions and the denominator specifies the total number of predictions.



\subsection{Experimental results on TUM-GAID} \label{subsec:expers}
%
%
In this section, we work with TUM-GAID dataset to evaluate the impact of CNN architectures in automatic extraction of gait signatures from diverse modalities, studying which one is the most convenient for the different scenarios.
Afterwards, we evaluate the impact of combining gait signatures from different modalities for people identification.
Finally, we compare our results with the \SOTA ones.

\subsubsection{Single modality evaluation}\label{subsec:experfeats}
In this experiment, we evaluate the individual contribution of each input modality (\ie gray pixels, optical flow and depth maps) and each architecture (\ie 2D, 3D and ResNet) for extracting discriminative gait signatures. We apply the metrics explained above (Sec.~\ref{subsec:metrics}). Note that, as this dataset only contains a single viewpoint, ResNet models tend to overfit due to the lack of variability in the training data. Therefore, we use ResNet-A (see Sec.~\ref{subsec:cnnarch} for more details) which is a shallower model designed for smaller datasets than the default ResNet models.
Tab~\ref{tab:tumSingle} summarizes the identification results obtained on TUM-GAID with each modality: \textit{Gray}, \textit{OF} and \textit{Depth}.
Each column contains the results for rank-1 (R1) and rank-5 (R5) for each scenario. The last column `\textit{AVG}' is the average of each case (temporal and non temporal) weighted by the number of classes. %

Focusing on the average results, we can see that the best R1 results are obtained by the 3D-CNN using  \textit{OF} modality. This clearly shows the importance of the temporal information captured by the 3D convolutions and the relevance of the motion information provided by the \textit{OF}. Regarding the best R5 result, it is obtained by the 2D-CNN using again the \textit{OF} modality. For this metric, the 2D-CNN produces better results than the others architectures showing the robustness of this model, \ie it is able to include the correct identity in the top-5 predictions more times than the other models. Finally, note that the two strategies employed for obtaining the identity at video level, \textit{SM-Vote} and \textit{SM-Prod}, offer similar performance. However, \textit{SM-Prod} seems to work slightly better in terms of average R1.

\begin{table}[tbh]
\small
\caption{\textbf{Modality selection on TUM-GAID. } Percentage of correct recognition by using \textit{rank-1} (R1) and \textit{rank-5} (R5) metrics. Each row corresponds to a different classifier and modality. Each column corresponds to a different scenario. Best average results are marked in bold.} 
\label{tab:tumSingle}
\begin{center}
\setlength{\tabcolsep}{0.3em} 
\renewcommand{\arraystretch}{1.3} 
\begin{tabular}{c c c|c c c c c c | c c c c c c|c c|}
\cline{4-17}
 & & & \multicolumn{2}{c}{\textit{N}} & \multicolumn{2}{c}{\textit{B}} & \multicolumn{2}{c|}{\textit{S}} & \multicolumn{2}{c}{\textit{TN}} & \multicolumn{2}{c}{\textit{TB}} & \multicolumn{2}{c|}{\textit{TS}} & \multicolumn{2}{c|}{\textit{AVG}}\\
  \cline{4-17}
 & & & R1 & R5 & R1 & R5 & R1 & R5 & R1 & R5 & R1 & R5 & R1 & R5 & R1 & R5 \\
 \hline
  \multicolumn{1}{|c|}{\multirow{6}{*}{\rotatebox{90}{\scriptsize 2D-CNN}}} & \multicolumn{1}{c|}{\multirow{2}{*}{\rotatebox{90}{Gray}}} & \emph{SM-Vote} & 99.4 & 100 & 99.0 & 99.7 & 98.4 & 99.7 & 31.3 & 53.1 & 34.4 & 65.6 & 34.4 & 62.5 & 92.8 & 96.1 \\
  \multicolumn{1}{|c|}{}& \multicolumn{1}{c|}{} &\emph{SM-Prod} & 100 & 100 & 99.7 & 99.7 & 98.4 & 99.7 & 28.1 & 62.5 & 37.5 & 71.9 & 34.4 & 59.4 & 93.2 & 96.5 \\
  \cline{2-17}
  \multicolumn{1}{|c|}{} & \multicolumn{1}{c|}{\multirow{2}{*}{\rotatebox{90}{OF}}} & \emph{SM-Vote} & 99.4 & 100 & 97.4 & 100 & 96.4 & 99.4 & 53.1 & 96.9 & 43.8 & 87.5 & 56.3 & 93.8 & 93.4 & \textbf{99.1} \\
  \multicolumn{1}{|c|}{} & \multicolumn{1}{c|}{} & \emph{SM-Prod} & 99.4 & 100 & 97.7 & 100 & 96.1 & 99.4 & 56.3 & 87.5 & 43.8 & 84.4 & 59.4 & 90.6 & 93.6 & 98.7 \\
  \cline{2-17}
  \multicolumn{1}{|c|}{} & \multicolumn{1}{c|}{\multirow{2}{*}{\rotatebox{90}{Depth}}} & \emph{SM-Vote} & 98.4 & 100 & 65.8 & 90.0 & 96.8 & 99.7 & 34.4 & 93.8 & 34.4 & 93.8 & 50.0 & 84.4 & 82.6 & 96.0 \\
  \multicolumn{1}{|c|}{} & \multicolumn{1}{c|}{} &\emph{SM-Prod} & 98.7 & 100 & 66.1 & 90.7 & 96.8 & 99.7 & 43.8 & 90.6 & 40.6 & 87.5 & 46.8 & 84.4 & 83.1 & 95.9 \\
  \hline
  \hline
  \multicolumn{1}{|c|}{\multirow{6}{*}{\rotatebox{90}{\scriptsize 3D-CNN}}} & \multicolumn{1}{c|}{\multirow{2}{*}{\rotatebox{90}{Gray}}} & \emph{SM-Vote} & 99.7 & 100 & 98.4 & 99.7 & 96.8 & 99 & 21.9 & 50 & 21.9 & 46.9 & 12.5 & 43.8 & 90.9 & 94.6 \\
  \multicolumn{1}{|c|}{} & \multicolumn{1}{c|}{} &\emph{SM-Prod} & 97.7 & 97.7 & 93.9 & 94.2 & 91.3 & 91.6 & 18.8 & 37.5 & 21.9 & 37.5 & 12.5 & 31.3 & 87.1 & 89 \\
  \cline{2-17}
  \multicolumn{1}{|c|}{} & \multicolumn{1}{c|}{\multirow{2}{*}{\rotatebox{90}{OF}}} & \emph{SM-Vote} & 99 & 99.4 & 95.5 & 99.7 & 94.2 & 98.1 & 65.6 & 90.6 & 65.6 & 93.8 & 59.4 & 87.5 & 93.2 & 98.3 \\
  \multicolumn{1}{|c|}{} & \multicolumn{1}{c|}{} &\emph{SM-Prod} & 98.7 & 99.7 & 97.1 & 99.4 & 94.5 & 98.7 & 71.9 & 87.5 & 68.8 & 87.5 & 65.6 & 84.4 & \textbf{94.1} & 98.1 \\
  \cline{2-17}
  \multicolumn{1}{|c|}{} & \multicolumn{1}{c|}{\multirow{2}{*}{\rotatebox{90}{Depth}}} & \emph{SM-Vote} & 97.7 & 98.4 & 84.2 & 96.5 & 96.8 & 99.4 & 68.8 & 100 & 50 & 100 & 75 & 100 & 90.3 & 98.3 \\
  \multicolumn{1}{|c|}{} & \multicolumn{1}{c|}{} & \emph{SM-Prod} & 98.4 & 100 & 86.8 & 96.1 & 97.4 & 99.4 & 62 & 87.4 & 53.1 & 96.9 & 78.1 & 100 & 91.4 & 98.2 \\
  \hline
  \hline
  \multicolumn{1}{|c|}{\multirow{6}{*}{\rotatebox{90}{\scriptsize RESNET-A}}} & \multicolumn{1}{c|}{\multirow{2}{*}{\rotatebox{90}{Gray}}} & \emph{SM-Vote} & 99.4 & 100 & 95.8 & 99.4 & 96.1 & 99 & 25 & 62.5 & 34.4 & 98.8 & 25 & 59.4 & 90.6 & 96.1 \\
  \multicolumn{1}{|c|}{}& \multicolumn{1}{c|}{} &\emph{SM-Prod} & 99 & 100 & 96.5 & 99.4 & 95.5 & 99 & 28.1 & 56.3 & 34.4 & 68.8 & 25 & 56.3 & 90.7 & 95.8 \\
  \cline{2-17} 
  \multicolumn{1}{|c|}{} & \multicolumn{1}{c|}{\multirow{2}{*}{\rotatebox{90}{OF}}} & \emph{SM-Vote} & 94.5 & 99.7 & 81 & 98.4 & 85.1 & 97.7 & 34.4 & 93.8 & 34.4 & 90.6 & 37.5 & 93.8 & 82.1 & 98.1 \\
  \multicolumn{1}{|c|}{} & \multicolumn{1}{c|}{} & \emph{SM-Prod} & 95.2 & 99.4 & 81 & 98.7 & 86.1 & 97.7 & 34.4 & 96.7 & 40.6 & 93.8 & 43.8 & 93.8 & 83 & 98.2 \\
  \cline{2-17}
  \multicolumn{1}{|c|}{} & \multicolumn{1}{c|}{\multirow{2}{*}{\rotatebox{90}{Depth}}} & \emph{SM-Vote} & 77.7 & 99.4 & 60 & 91.6 & 71.8 & 97.4 & 56.3 & 81.3 & 37.5 & 81.3 & 46.9 & 90.6 & 67.7 & 95\\
  \multicolumn{1}{|c|}{} & \multicolumn{1}{c|}{} & \emph{SM-Prod} & 77.1 & 98.4 & 60 & 91.3 & 70.9 & 96.1 & 56.3 & 78.1 & 34.4 & 81.3 & 46.9 & 87.5 & 67.1 & 94.1 \\
  \hline
\end{tabular}
\end{center}
\end{table}

\begin{figure}[tbh!]
\begin{center}
   \includegraphics[width=0.99\linewidth]{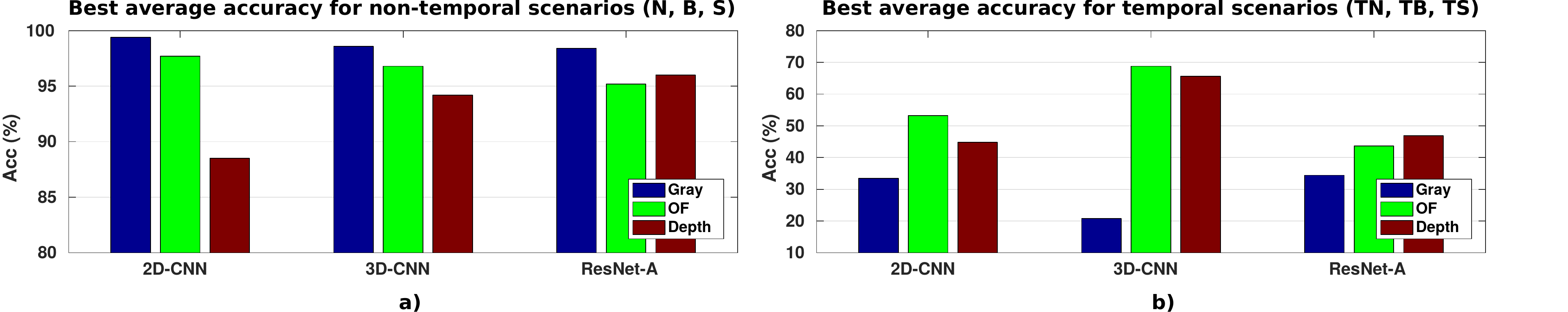}
\end{center}
\caption{\textbf{Best average R1 accuracy}. 
\textbf{a)} non-temporal scenarios (N, B, S)
\textbf{b)} temporal scenarios (TN, TB, TS)
}
\label{fig:com_modalities}
\end{figure}

In order to compare the performance of the different models and modalities per scenario (\ie temporal and non-temporal), we summarize the most important results of Tab.~\ref{tab:tumSingle} in Fig.~\ref{fig:com_modalities}. It shows the best R1 average performance for non-temporal and temporal scenarios per modality. If we focus on the non-temporal scenarios (\textit{N}, \textit{B} and \textit{S}), we can see gait signatures based on \textit{Gray} are able to outperform the results obtained with \textit{OF}. In our opinion, appearance models have such a good performance due to the low appearance variability of this scenario. On the other hand, if we focus on the temporal scenarios (\textit{TN}, \textit{TB} and \textit{TS}), the worst results are obtained with \textit{Gray}. These results evidence the weakness of appearance models under conditions with high variability between training and test samples (like our temporal experiment). However, \textit{OF} models have a better sturdiness against appearance changes on the inputs.
With regard to the type of architecture, the behaviour of all of them is very similar on the non-temporal scenarios, where the best results are obtained by the 2D-CNN using \textit{Gray} as input, followed by the 3D-CNN with \textit{Gray}. However, for the temporal scenarios, 3D-CNN offers its best results in using either \textit{OF} or \textit{Depth}, whereas 2D-CNN and ResNet work better with \textit{Gray}.
Considering the average accuracy over all the scenarios, 2D-CNN works better with \textit{Gray}, and 3D-CNN with both \textit{OF} and \textit{Depth}.

Thus, we can conclude that the best architectures for this task are 2D-CNN and 3D-CNN, and the best modalities are \textit{Gray} for the non-temporal scenario and \textit{OF} for the temporal scenario.

\begin{table}[htb]
\small
\caption{\textbf{Fusion position for all networks.} Percentage of correct rank-1 (R1) recognition for all scenarios and fusion modalities. Each row corresponds to a different architecture and each column corresponds to a different fusion position. Best results are marked in bold.} 
\label{tab:tumFusionPos}
\begin{center}
\setlength{\tabcolsep}{0.4em} 
\renewcommand{\arraystretch}{0.9} 
\begin{tabular}{c|c c c c c|c c c c c|c c c c c|}
\cline{2-16}
& \multicolumn{5}{c|}{OF-Gray} & \multicolumn{5}{c|}{OF-Depth} & \multicolumn{5}{c|}{All} \\
\cline{2-16}
& P1 & P2 & P3 & P4 & P5 & P1 & P2 & P3 & P4 & P5 & P1 & P2 & P3 & P4 & P5 \\
\hline
\multicolumn{1}{|c|}{2D-CNN} & 90.7 & 91.9 & 91.8 & \textbf{93.9} & 93.8 & 81.8 & 92.7 & 91.9 & 92.2 & \textbf{93.1} & 86.4 & 90.1 & 92.8 & 93.5 & \textbf{93.9} \\
\multicolumn{1}{|c|}{3D-CNN} & 92.3 & 91.3 & 89.4 & 92.7 & \textbf{92.9} & 86.6 & 78.6 & 71.8 & 89.0 & \textbf{89.2} & 87.9 & 82.5 & 87.3 & 92.2 & \textbf{92.7} \\
\multicolumn{1}{|c|}{ResNet} & 86.8 & 89.4 & 89.3 & 92.4 & \textbf{92.6} & 83.2 & 83.1 & 85.5 & \textbf{90.4} & 90.0 & 84.7 & 87.8 & 88.2 & \textbf{92.9} & 90.6 \\
\hline 
\end{tabular}
\end{center}
\end{table}

\subsubsection{Multiple modality evaluation}\label{subsec:experfusion}
As we can use three types of modalities from TUM-GAID, we study here the benefits of fusing information from different modalities. 
In this section, we compare the baseline obtained by each single modality with our two late fusion approaches, which are performed at score level, and our early fusion approach, which is performed at gait signature level. For more details about our fusion approaches (early and late methods), please see Sec.~\ref{subsec:multiclasif}. Since there are many combinations of modalities, our experiments focus just on the combinations that include optical flow due to the sturdiness of this modality under all walking conditions.

Note that, in this section, the data of 
Tab.~\ref{tab:tumSingle} are used as baseline case, concretely, data obtained with the \textit{product of softmax vectors} on each modality.

Focusing on the early fusion approach, we have studied the impact of fusion on network performance attending to the position where fusion is applied in the architecture.  
Thus, Tab.~\ref{tab:tumFusionPos} summarizes the average results of three trials per network for five different fusion positions. For 2D-CNN and 3D-CNN, the positions P1 to P4 refer to  the convolutional layers and P5 allude to the fully-connected layer. For RestNet, P1 to P4 correspond to the last convolutional layer of each residual block and P5 is the average pooling layer. Focusing on the results, we can see that for 2D-CNN and 3D-CNN, the best results are obtained in the last fusion position. For ResNet, P4 and P5 achieve similar results in most of the cases, so we are going to focus on P5 since this layer produces an uni-dimensional feature vector which does not requires any further pre-processing in order to be used by our fusion approaches. Note that in the following tables, the early fusion results are selected from the best position and the best of the three trials.

Now, we analyse the fusion results within each type of architecture.
The results of Tab.~\ref{tab:tumFusion2D} correspond to 2D-CNN and indicate that, in general, the best option is to combine all three modalities for all fusion methods except for \textit{SM Prod} where it is better to use only \textit{OF} and \textit{Gray}. Note that for \textit{Weighted Sum}, we have used the weights $0.4, 0.3$ and $0.3$ for \textit{OF}, \textit{Gray} and \textit{Depth}, respectively, when we fuse all modalities. 
In the case of only two modalities, we use fusion weights $0.6$ and $0.4$ for \textit{OF} and the other modality, respectively. According to the results, in the non temporal scenarios we only improve the results with respect to the single modalities in scenario \textit{S}, while in the other scenarios we obtain similar results (since they are higher than $99.7\%$, the improvement is very difficult). 
Regarding the fusion strategy, the proposed \textit{Early} fusion CNN provides on average the best results showing that it is better to combine the information from multiple modalities during the training process. By this way, the parameters of the network are able to learn the best combination that maximizes the accuracy.

\begin{table}[tbh]
\small
\caption{\textbf{Fusion strategies in TUM-GAID with 2D-CNN.} Percentage of correct rank-1 (R1) recognition for different modalities and fusion methods. Each row corresponds to a different fusion strategy. Best results are marked in bold.} 
\label{tab:tumFusion2D}
\begin{center}
\renewcommand{\arraystretch}{0.9} 
\begin{tabular}{|c|l|c c c|c c c|c|}
\hline
Fusion & Modalities & \textit{N} & \textit{B} & \textit{S} & \textit{TN} & \textit{TB} & \textit{TS} & \textit{AVG}\\
 \hline
 \multirow{3}{*}{Single} & Gray & 100 & 99.7 & 98.4 & 28.1 & 37.5 & 34.4 & 93.2 \\
 & OF & 99.4 & 97.7 & 96.1 & 56.3 & 43.8 & 59.4 & 93.6 \\
 & Depth & 98.7 & 66.1 & 96.8 & 43.8 & 40.6 & 46.8 & 83.1 \\
 \hline
 \hline 
    \multirow{3}{*}{SM Prod} & OF-Gray & 99.7 & 99.7 & 99.0 & 40.6 & 37.5 & 53.1 & 94.3 \\
  & OF-Depth & 92.9 & 88.1 & 98.7 & 59.4 & 40.6 & 46.9 & 89.1 \\
  & All & 92.9 & 90.0 & 99.0 & 56.3 & 56.3 & 50.0 & 90.2 \\
  \hline
  \multirow{3}{*}{W. Sum} & OF-Gray & 99.4 & 98.4 & 98.7 & 50.0 & 34.4 & 53.1 & 93.9 \\
  & OF-Depth & 97.7 & 93.9 & 99.0 & 53.1 & 43.8 & 59.4 & 92.7 \\
  & All & 99.0 & 98.1 & 99.7 & 50.0 & 34.4 & 53.1 & 94.0 \\
  \hline
  \multirow{3}{*}{Early} & OF-Gray & 99.4 & 98.7 & 97.7 & 56.3 & 43.8 & 43.8 & 93.9  \\
  & OF-Depth & 99.0 & 96.1 & 96.4 & 53.1 & 53.1 & 46.9 & 92.9 \\
  & All & 99.4 & 98.4 & 98.7 & 56.3 & 53.1 & 46.9 & \textbf{94.5} \\
  \hline 
\end{tabular}
\end{center}
\end{table}

Focusing on the results obtained with the 3D-CNN  (Tab.~\ref{tab:tumFusion3D}), the best average accuracy is reported by the combination of all modalities by \textit{W Sum}. However, it is only slightly better than the best result obtained by using only \textit{OF}.
Due to the low accuracy obtained with \textit{Gray}, combining it with other features worsen the fused results.

\begin{table}[tbh]
\small
\caption{\textbf{Fusion strategies in TUM-GAID with 3D-CNN.} Percentage of correct rank-1 (R1) recognition for different modalities and fusion methods. Each row corresponds to a different fusion strategy. Best results are marked in bold.} 
\label{tab:tumFusion3D}
\begin{center}
\renewcommand{\arraystretch}{0.9} 
\begin{tabular}{|c|l|c c c|c c c|c|}
\hline
Fusion & Modalities & \textit{N} & \textit{B} & \textit{S} & \textit{TN} & \textit{TB} & \textit{TS} & \textit{AVG}\\
 \hline
 \multirow{3}{*}{Single} & Gray & 97.7 & 93.9 & 91.3 & 18.8 & 21.9 & 12.5 & 87.1 \\
 & OF & 98.7 & 97.1 & 94.5 & 71.9 & 68.8 & 65.6 & 94.1 \\
 & Depth & 98.4 & 86.8 & 97.4 & 62 & 53.1 & 78.1 & 91.4 \\
 \hline
 \hline 
    \multirow{3}{*}{SM Prod} & OF-Gray & 93.5 & 84.8 & 83.5 & 12.5 & 12.5 & 15.6 & 80.4 \\
  & OF-Depth & 92.2 & 97.4 & 96.8 & 78.1 & 62.5 & 15.6 & 91.4 \\
  & All & 78.4 & 84.2 & 83.5 & 12.5 & 21.9 & 12.5 & 75.8 \\
  \hline
  \multirow{3}{*}{W. Sum} & OF-Gray & 97.4 & 98.1 & 96.1 & 71.9 & 50 & 53.1 & 93.6 \\
  & OF-Depth & 95.5 & 96.5 & 96.8 & 65.6 & 68.8 & 53.1 & 93.1 \\
  & All & 96.8 & 98.4 & 97.1 & 65.6 & 65.6 & 59.4 & \textbf{94.3} \\
  \hline
  \multirow{3}{*}{Early} & OF-Gray & 99.4 & 96.8 & 94.5 & 62.5 & 50 & 56.3 & 93.1 \\
  & OF-Depth & 84.8 & 97.4 & 97.4 & 71.9 & 68.8 & 71.9 & 91.1\\
  & All & 99.7 & 98.7 & 97.7 & 34.4 & 25 & 31.3 & 92.3 \\
  \hline 
\end{tabular}
\end{center}
\end{table}

Finally, the ResNet architecture 
(see Tab.~\ref{tab:tumFusionRES}) shows unexpected low fusion results. It may indicate that the probability distribution on the classes obtained at the softmax layer does not show clearly defined maxima, and small changes in those values cause important changes in the final classes. However, \textit{Early Fusion} improves the results on average for the combination \textit{OF} and \textit{Gray}, what indicates that adding more inputs to the training process can be beneficial to avoid local minima.

In summary, by using multimodal information the recognition accuracy improves $0.9\%$ with respect to the best single modality (i.e. \textit{OF}).

\begin{table}[tbh]
\small
\caption{\textbf{Fusion strategies in TUM-GAID with ResNet.} Percentage of correct rank-1 (R1) recognition for different modalities and fusion methods. Each row corresponds to a different fusion strategy. Best average results are marked in bold.} 
\label{tab:tumFusionRES}
\begin{center}
\renewcommand{\arraystretch}{0.9} 
\begin{tabular}{|c|l|c c c|c c c|c|}
\hline
Fusion & Modalities & \textit{N} & \textit{B} & \textit{S} & \textit{TN} & \textit{TB} & \textit{TS} & \textit{AVG}\\
 \hline
 \multirow{3}{*}{Single} & Gray & 99.0 & 96.5 & 95.5 & 28.1 & 34.4 & 25.0 & 90.7 \\
 & OF & 95.2 & 81.0 & 86.1 & 37.5 & 40.6 & 43.8 & 83.1 \\
 & Depth & 77.1 & 60.0 & 71.0 & 56.3 & 34.4 & 46.9 & 67.2 \\
 \hline
 \hline 
    \multirow{3}{*}{SM Prod} & OF-Gray & 84.8 & 77.7 & 79.3 & 46.9 & 40.6 & 50 & 77.3\\
  & OF-Depth & 71.2 & 63.6 & 69.6 & 53.1 & 37.5 & 53.1 & 66.2\\
  & All & 79.9 & 80.7 & 81.9 & 56.3 & 34.4 & 56.3 & 77.9\\
  \hline
  \multirow{3}{*}{W. Sum} & OF-Gray & 72.8 & 60.7 & 64.4 & 31.3 & 31.3 & 40.6 & 63\\
  & OF-Depth & 68.3 & 53.9 & 62.5 & 37.5 & 46.9 & 56.3 & 60.2\\
  & All & 72.5 & 60 & 64.7 & 31.3 & 28.1 & 46.9 & 62.9\\
  \hline
  \multirow{3}{*}{Early-RES} & OF-Gray & 99.4 & 94.8 & 97.7 & 40.6 & 34.4 & 43.8 & \textbf{91.9} \\
  & OF-Depth & 95.8 & 93.2 & 96.1 & 40.6 & 37.5 & 43.8 & 89.9 \\
  & All & 80.3 & 87.1 & 88.4 & 40.6 & 50 & 50 & 81.7 \\
  \hline 
\end{tabular}
\end{center}
\end{table}


\subsubsection{State-of-the-art on TUM-GAID}\label{subsec:sota}
In Tab.~\ref{tab:sotaTUM}, we compare our results with the \SOTA in TUM-GAID under all modalities previously employed (\textit{Gray}, \textit{OF}, \textit{Depth} and \textit{Fusion}). First of all, we would like to remark that our approach uses a resolution of $80\times 60$ while the rest of the methods use $640\times 480$. Therefore, our method uses $64$ times less information. If we focus on the visual modality (\textit{Gray} in our case), we can see that our method outperforms previous results in non temporal scenarios establishing a new \SOTA. On the other hand, in the temporal scenarios we have lower results than the other methods due to the high variability in visual information.
Then, if we focus on \textit{OF}, we can see that the best results are obtained by PFM~\cite{castro2016ijprai} with a resolution of $640\times 480$. Nevertheless, if we apply PFM with a resolution of $80\times 60$, its results worsen dramatically and our CNN is able to outperform it in all scenarios. If we compare our CNN with other deep learning approaches presented in the literature, only MTaskCNN-7NN~\cite{marin2017icip} is able to improve our approach. This model has been trained in a multi-task fashion so, during training, there are more information available to optimize the network parameters. If we focus on the other deep learning approaches, we can see that we obtain similar results (only a $0.2\%$ lower) on average but, we obtain the \SOTA for temporal scenario.
In \textit{Depth} modality, we can see that our method obtains better results than other methods, which use full resolution frames, in all cases except \textit{N}. Nevertheless, on average, we are able to obtain more than a $10\%$ of improvement.
Finally, if we fuse information from all modalities with a CNN, the average score achieved by both scenarios (temporal and non-temporal) beats all the methods shown in Tab.~\ref{tab:sotaTUM} with the exception of PFM (640x480)~\cite{castro2016ijprai} and MTaskCNN-7NN~\cite{marin2017icip}, where we are $1.5\%$ and $1.1\%$ below, respectively. However, if we apply the same 7NN approach as in~\cite{marin2017icip}, and we fuse the probabilities obtained, we set a new \SOTA ($96.5\%$ vs $96.0\%$) for all scenarios with our 3D-CNN-7NN-All using Softmax Product as fusion.

\begin{table}[tbh] 
\small
\centering
\caption{\textbf{State-of-the-art on TUM GAID}. Percentage of correct rank-1 (R1) recognition on TUM-GAID for diverse methods published in the literature. Bottom rows of each modality correspond to our proposal, where instead of using video frames at $640\times 480$, a resolution of $80\times 60$ is used. Each column corresponds to a different scenario. Best results are marked in bold. (See main text for further details). }
\label{tab:sotaTUM}
\small
\setlength{\tabcolsep}{0.25em} 
\renewcommand{\arraystretch}{0.9} 
\begin{tabular}{|c|c|c|ccc|c||ccc|c||c|}
\hline 
\textit{Modality} & \textit{Input Size} & \textit{Method} & \textit{N} & \textit{B} & \textit{S} & \textit{Avg} & \textit{TN} & \textit{TB} & \textit{TS} & \textit{Avg} & \textit{Global Avg}\\ 
\hline 
\multirow{8}{*}{\rotatebox{90}{Visual Data}} & \multirow{6}{*}{\rotatebox{90}{$640\times 480$}} &SDL~\cite{zeng2014pr} & - & - & - & - & 96.9 & - & - &  - & -\\ 
&&GEI~\cite{hofmann2014tumgaid} & 99.4 & 27.1 & 52.6 & 59.7 & 44.0 & 6.0 & 9.0 & 19.7 & 56.0\\
&&SEIM~\cite{whytock2014jmiv} & 99.0 & 18.4 & 96.1 & 71.2 & 15.6 & 3.1 & 28.1 & 15.6 & 66.0\\
&&GVI~\cite{whytock2014jmiv} & 99.0 & 47.7 & 94.5 & 80.4 & 62.5 & 15.6 & 62.5 & 46.9 & 77.3\\
&&SVIM~\cite{whytock2014jmiv} & 98.4 & 64.2 & 91.6 & 84.7 & 65.6 & 31.3 & 50.0 & 49.0 & 81.4\\
&&RSM~\cite{guan2013icb} & 100 & 79.0 & 97.0 & 92.0 & 58.0 & 38.0 & 57.0 & 51.0 & 88.2\\
\cline{2-12}
&Gray $80\times 60$&2D-CNN-SMP \textbf{(ours)} & 100 & 99.7 & 98.4 & 99.4 & 28.1 & 37.5 & 34.4 & 33.3 & 93.2\\
\hline
\multirow{6}{*}{OF} & $640\times 480$ & PFM~\cite{castro2016ijprai} & 99.7 & 99.0 & 99.0 & 99.2 & 78.1 & 62.0 & 54.9 & 65.0 & 96.0\\
\cline{2-12}
&\multirow{5}{*}{$80\times 60$}&PFM~\cite{castro2016ijprai} & 75.8 & 70.3 & 32.3 & 59.5 & 50.0 & 40.6 & 25.0 & 38.5 & 57.5\\
& & OF-CNN-NN~\cite{castro2017iwann} & 99.7 & 98.1 & 95.8 & 97.9 & 62.5 & 56.3 & 59.4 & 59.4 & 94.3 \\
& & OF-ResNet-B~\cite{castro2017biosig} & 99 & 95.5 & 97.4 & 97.3 & 65.6 & 62.5 & 68.8 & 65.6 & 94.3 \\
& & MTaskCNN-7NN~\cite{marin2017icip} & 99.7 & 97.4 & 99.7 & 98.9 & 59.4 & 62.5 & 68.8 & 63.6 & 95.6 \\
& &3D-CNN-SMP \textbf{(ours)} & 98.7 & 97.1 & 94.5 & 96.8 & 71.9 & 68.8 & 65.6 & \textbf{68.8} & 94.1\\
\hline
\multirow{2}{*}{Depth} & $640\times 480$ & DGHEI~\cite{hofmann2014tumgaid} & 99.0 & 40.3 & 96.1 & 78.5 & 50.0 & 0.0 & 44.0 & 31.3 & 74.1 \\
\cline{2-12}
&$80\times 60$&3D-CNN-SMP \textbf{(ours)} & 98.4 & 86.8 & 97.4 & 94.2 & 62.0 & 53.1 & 78.1 & 64.4 & 91.4\\
\hline
\multirow{3}{*}{Fusion}& $640\times 480$ & DGHEI + GEI~\cite{hofmann2014tumgaid} & 99.4 & 51.3 & 94.8 & 81.8 & 66.0 & 3.0 & 50.0 & 39.7 & 77.9\\
\cline{2-12}
&\multirow{2}{*}{$80\times 60$} & 2D-CNN-All \textbf{(ours)} & 99.4 & 98.4 & 98.7 & 98.8 & 56.3 & 53.1 & 46.9 & 52.1 &94.5 \\
&& 3D-CNN-7NN-All \textbf{(ours)} & 100 & 99.4 & 99.4 & \textbf{99.6} & 75 & 62.5 & 62.5 & 66.7 & \textbf{96.5} \\
\hline
\end{tabular}
\end{table}

\subsection{Experimental results on CASIA-B}\label{subsec:experCASIAB}
We focus here on CASIA-B dataset, which offers different covariate factors and multiple viewpoints. Note that, for the sake of comparison with other methods, 
we train our models with all cameras and we test them both on the $90\degree$ camera, as done in the \SOTA approaches~\cite{Wu2016pami, castro2016ijprai}, and on all cameras like in~\cite{Wu2016pami}.

\subsubsection{Single modality evaluation}\label{subsec:experfeatsCASIAB}
%
As this dataset contains eleven viewpoints, ResNet models have enough variability in the training data. Therefore, we use ResNet-B (see Sec.~\ref{subsec:cnnarch} for more details) which is deeper than ResNet-A.
Tabs.~\ref{tab:casiaSingleGray} and~\ref{tab:casiaSingleAll} summarize the identification results obtained on CASIA-B $90^o$ and multicamera setup, respectively, for each modality: \textit{Gray} and \textit{OF}. Note that this dataset does not provide depth information.
R1 and R5 columns contain the results for rank-1 (R1) and rank-5 (R5) for each scenario. The last column `\textit{AVG}' is the average of all scenarios. The results at sequence level are obtained by multiplying the scores of the softmax layer. Note that as in CASIA-B there is no training partition to build the model, we have split the dataset into a training set composed of the first 74 subjects and a test set composed of the 50 remaining subjects, following the indications in~\cite{Wu2016pami}. During the training process, all viewpoints and training samples are used. 

\begin{table}[tbh]
\small
\caption{\textbf{Modality selection on CASIA-B $\textbf{90\degree}$: \textit{Gray} and \textit{OF} modalities.} Percentage of correct recognition by using \textit{rank-1} (R1) and \textit{rank-5} (R5) metrics. Each row corresponds to a different classifier and modality, grouped by architecture. Each column corresponds to a different scenario. 
Best average results are marked in bold.} 
\label{tab:casiaSingleGray}
\begin{center}
\setlength{\tabcolsep}{0.3em} 
\renewcommand{\arraystretch}{0.9} 
\begin{tabular}{|c|l|l | c c c c c c|c c|}
 \cline{4-11} 
 \multicolumn{1}{c}{}&\multicolumn{1}{c}{}&  & \multicolumn{2}{c}{\textit{nm}} & \multicolumn{2}{c}{\textit{bg}} & \multicolumn{2}{c|}{\textit{cl}} & \multicolumn{2}{c|}{\textit{AVG}}\\
 \cline{4-11}
 \multicolumn{1}{c}{}&\multicolumn{1}{c}{}& & R1 & R5 & R1 & R5 & R1 & R5 & R1 & R5 \\
 \hline
  \multirow{6}{*}{\rotatebox{90}{Gray}} & \multirow{2}{*}{\rotatebox{90}{2D}} &\emph{SM-Vote} & 91 & 98 & 82 & 95 & 37 & 82 & 70 & 91.7  \\
  &&\emph{SM-Prod} & 92 & 100 & 85 & 98 & 45 & 90 & 74 & 96   \\
 \cline{2-11} 
  &\multirow{2}{*}{\rotatebox{90}{3D}} &\emph{SM-Vote} & 72 & 93 & 69 & 87 & 33 & 76 & 58 & 85.3 \\
  &&\emph{SM-Prod} & 81 & 92 & 73 & 90 & 45 & 77 & 66.3 & 86.3 \\
  \cline{2-11} 
  &\multirow{2}{*}{\rotatebox{90}{RES}} &\emph{SM-Vote} & 94 & 100 & 89 & 98 & 42 & 83 & 75 & 93.7 \\
  &&\emph{SM-Prod} & 96 & 100 & 91 & 98 & 46 & 98 & \textbf{77.7} & \textbf{98.7} \\
  \hline 
 \hline
  \multirow{6}{*}{\rotatebox{90}{OF}} & \multirow{2}{*}{\rotatebox{90}{2D}} &\emph{SM-Vote} & 99 & 99 & 76 & 90 & 28 & 51 & 67.7 & 80  \\
  &&\emph{SM-Prod} & 99 & 99 & 78 & 93 & 27 & 62 & 68 & 84.7   \\
  \cline{2-11} 
  &\multirow{2}{*}{\rotatebox{90}{3D}} &\emph{SM-Vote} & 98 & 99 & 86 & 99 & 37 & 70 & 73.7 & 89.3  \\
  &&\emph{SM-Prod} & 98 & 100 & 88 & 98 & 36 & 67 & 74 & 88.3  \\
  \cline{2-11} 
  &\multirow{2}{*}{\rotatebox{90}{RES}} &\emph{SM-Vote} & 94 & 100 & 83 & 98 & 47 & 73 & 74.7 & 90.3 \\
  &&\emph{SM-Prod} & 93 & 100 & 85 & 98 & 46 & 71 & 74.7 & 89.7 \\
  \hline   
\end{tabular}
\end{center}
\end{table}

\begin{table}[tbh]
\small
\caption{\textbf{Modality selection on CASIA-B all cameras: \textit{Gray} and \textit{OF} modalities.} Percentage of correct recognition by using \textit{rank-1} (R1) and \textit{rank-5} (R5) metrics. Each row corresponds to a different classifier and modality, grouped by architecture. Each column corresponds to a different scenario. 
Best average results are marked in bold.} 
\label{tab:casiaSingleAll}
\begin{center}
\setlength{\tabcolsep}{0.3em} 
\renewcommand{\arraystretch}{0.9} 
\begin{tabular}{|c|l|l | c c c c c c|c c|}
 \cline{4-11} 
 \multicolumn{1}{c}{}&\multicolumn{1}{c}{}&  & \multicolumn{2}{c}{\textit{nm}} & \multicolumn{2}{c}{\textit{bg}} & \multicolumn{2}{c|}{\textit{cl}} & \multicolumn{2}{c|}{\textit{AVG}}\\
 \cline{4-11}
 \multicolumn{1}{c}{}&\multicolumn{1}{c}{}& & R1 & R5 & R1 & R5 & R1 & R5 & R1 & R5 \\
 \hline
  \multirow{6}{*}{\rotatebox{90}{Gray}} & \multirow{2}{*}{\rotatebox{90}{2D}} &\emph{SM-Vote} & 97.1 & 99.6 & 89.8 & 98.5 & 38.4 & 78.7 & 75.1 & 92.3  \\
  &&\emph{SM-Prod} & 98.3 & 100 & 91.5 & 99.2 & 40.4 & 77.2 & 76.7 & 92.1   \\
 \cline{2-11} 
  &\multirow{2}{*}{\rotatebox{90}{3D}} &\emph{SM-Vote} & 92.7 & 98.8 & 87.1 & 96.3 & 43.9 & 72.8 & 74.6 & 89.3 \\
  &&\emph{SM-Prod} & 96.1 & 99.1 & 88.7 & 96.9 & 47.0 & 71.7 & 77.3 & 89.2 \\
  \cline{2-11} 
  &\multirow{2}{*}{\rotatebox{90}{RES}} &\emph{SM-Vote} & 99.2 & 100 & 91.6 & 99.4 & 44.1 & 78.4 & 78.3 & 92.6 \\
  &&\emph{SM-Prod} & 99.5 & 100 & 92.3 & 99.6 & 45.6 & 78.6 & \textbf{79.1} & 92.7 \\
  \hline 
 \hline
  \multirow{6}{*}{\rotatebox{90}{OF}} & \multirow{2}{*}{\rotatebox{90}{2D}} &\emph{SM-Vote} & 97.6 & 99.7 & 82.4 & 95.5 & 38.3 & 63.9 & 72.8 & 86.4  \\
  &&\emph{SM-Prod} & 97.6 & 99.8 & 83.4 & 96.3 & 37.9 & 68.5 & 73.0 & 88.2   \\
  \cline{2-11} 
  &\multirow{2}{*}{\rotatebox{90}{3D}} &\emph{SM-Vote} & 98.2 & 99.8 & 89.7 & 99.1 & 45.0 & 78.2 & 77.6 & 92.4  \\
  &&\emph{SM-Prod} & 98.4 & 99.9 & 90.5 & 98.8 & 45.1 & 76.7 & 78.0 & 91.8  \\
  \cline{2-11} 
  &\multirow{2}{*}{\rotatebox{90}{RES}} &\emph{SM-Vote} & 95.9 & 100 & 84.9 & 98.7 & 47.4 & 83.1 & 76.1 & \textbf{93.9} \\
  &&\emph{SM-Prod} & 96.0 & 100 & 86.7 & 98.5 & 48.0 & 79.9 & 76.9 & 92.8 \\
  \hline   
\end{tabular}
\end{center}
\end{table}

According to the results obtained for the $90\degree$ view-point and for all cameras, we can see that our model is able to identify people with a high accuracy in scenarios \textit{nm} and \textit{bg} while in scenario \textit{cl} we have lower precision due to the high appearance changes. If we focus on the modality used, on average, \textit{Gray} is the best option most of the time. In this dataset, with huge variations between points of view, the shape of the subject seems to be important and it helps to  classification. In scenario \textit{cl} our models experiment a huge decrease in accuracy mainly caused by the high variability of coats worn by the subjects. This can be seen in Fig.~\ref{fig:datasets} on the right part of the last row. In these pictures, the coat occludes the legs and if we add the fact that we have different kind of coats with different number of occurrences, our CNN is not able to learn good features for this scenario due to the high variability and low number of samples.

On the other hand, for this dataset, \textit{OF} seems that it is not able to find a good representation if the shape of the subject changes drastically. We think that this is because of the high variability in the appearance of the subjects seen from the different cameras used for training. Therefore, as the models receive different flow vectors, the training process cannot produce a view-independent model and the global performance decreases. For example, frontal-views produce vectors whose main movement is focused on Y-axis (there is no horizontal displacement of the subject) while lateral-views produce vectors whose movement is focused on X-axis.

If we analyze the average recognition percentage achieved by the different architectures, it is clear that ResNet-B obtains the best results for all modalities with the $90\degree$ camera. When all cameras are used for testing, ResNet-B obtains the best results for \textit{Gray} modality while 3D-CNN achieves the best results for \textit{OF}. That shows that ResNet is the most powerful model when data with enough variability is available. On the other hand, 3D-CNN obtains good results for the \textit{OF} modality, while 2D-CNN achieves good results for the \textit{Gray} modality.

Thus, we can conclude that the best architecture for this dataset is ResNet, and the best modalities are \textit{Gray} for R1 metric and \textit{OF} for R5 metric.

\subsubsection{Multiple modality evaluation}\label{subsec:experfusionCASIAB}
In this dataset only two modalities are available (\ie gray and OF) so fusion experiments are restricted to just both of them. 
As we have done for TUM-GAID dataset, we compare the baseline obtained by each single modality with our two late fusion approaches, which are performed at score level, and our early fusion approach, which is performed at gait signature level. Again, in this section, the data of Tabs.~\ref{tab:casiaSingleGray} and~\ref{tab:casiaSingleAll} are used as baseline case, concretely, data obtained with the \textit{product of softmax vectors} on each modality.

It can be observed in Tabs.~\ref{tab:casiaFusion} and~\ref{tab:casiaFusionAll} that the best method for fusing \textit{Gray} and \textit{OF} modalities is, on average, Softmax product followed by weighted sum (with weights $0.5$ and $0.5$ for \textit{Gray} and \textit{OF}, respectively). Focusing on the three architectures, ResNet obtains the best fusion results for the $90\degree$ camera while 2D-CNN achieves the best results using all cameras.

In this case, early fusion is not able to improve the single modality results. In our opinion, this is due to high variability between viewpoints. In addition, we have observed that the two branches of the network have different convergence speeds, hence the final features are not fused properly producing bad representations.

Thus, in this dataset, the late fusion (SM-Prod) improves the single modality results showing that both modalities are complementary and the combination of both boosts the performance of the models. Moreover, the boost is very significant in many cases with improvements higher than $9\%$ (\eg ResNet with all cameras).


\begin{table}[tbh]
\small
\caption{\textbf{Fusion strategies in CASIA-B 90\textdegree.} Percentage of correct rank-1 (R1) recognition with different fusion methods. Each row corresponds to a different fusion method, but the two top rows that correspond to the baseline cases. Best average results are marked in bold.} 
\label{tab:casiaFusion}
\begin{center}
\setlength{\tabcolsep}{0.3em} 
\renewcommand{\arraystretch}{0.9} 
\begin{tabular}{|l|c c c |c|| c c c |c|| c c c |c|}
\cline{2-13} 
 \multicolumn{1}{c|}{} & \multicolumn{4}{c||}{\textit{2D-CNN}} & \multicolumn{4}{c||}{\textit{3D-CNN}} & \multicolumn{4}{c|}{\textit{ResNet}} \\
 \cline{2-13} 
 \multicolumn{1}{c|}{}& \textit{nm} & \textit{bg} & \textit{cl} & \textit{AVG} & \textit{nm} & \textit{bg} & \textit{cl} & \textit{AVG} & \textit{nm} & \textit{bg} & \textit{cl} & \textit{AVG}\\
 \hline
  \emph{Gray} & 92 & 85 & 45 & 74 & 81 & 73 & 45 & 66.3 & 96 & 91 & 46 & 77.7 \\
   \emph{OF} & 99 & 78 & 27 & 68 & 98 & 88 & 36 & 74 & 93 & 85 & 46 & 74.7 \\
   \hline 
   \emph{SM-Prod} & 99 & 95 & 41 & \textbf{78.3} & 98 & 96 & 49 & \textbf{81} & 98 & 97 & 63 & \textbf{86} \\
   \emph{W. Sum} & 99 & 94 & 39 & 77.3 & 98 & 95 & 46 & 79.7 & 98 & 96 & 60 & 84.7 \\
   \emph{Early} & 83 & 61 & 26 & 56.7 & 76 & 74 & 46 & 65.3 & 67 & 63 & 38 & 56 \\
  \hline 
\end{tabular}
\end{center}
\end{table}

\begin{table}[tbh]
\small
\caption{\textbf{Fusion strategies in CASIA-B all cameras.} Percentage of correct rank-1 (R1) recognition with different fusion methods. Each row corresponds to a different fusion method, but the two top rows that correspond to the baseline cases. Best average results are marked in bold.} 
\label{tab:casiaFusionAll}
\begin{center}
\setlength{\tabcolsep}{0.3em} 
\renewcommand{\arraystretch}{0.9} 
\begin{tabular}{|l|c c c |c|| c c c |c|| c c c |c|}
\cline{2-13} 
 \multicolumn{1}{c|}{} & \multicolumn{4}{c||}{\textit{2D-CNN}} & \multicolumn{4}{c||}{\textit{3D-CNN}} & \multicolumn{4}{c|}{\textit{ResNet}} \\
 \cline{2-13} 
 \multicolumn{1}{c|}{}& \textit{nm} & \textit{bg} & \textit{cl} & \textit{AVG} & \textit{nm} & \textit{bg} & \textit{cl} & \textit{AVG} & \textit{nm} & \textit{bg} & \textit{cl} & \textit{AVG}\\
 \hline
  \emph{Gray} & 98.3 & 91.5 & 40.4 & 76.7 & 96.1 & 88.7 & 47.0 & 77.3 & 99.5 & 92.3 & 45.6 & 79.1 \\
   \emph{OF} & 97.6 & 83.4 & 37.9 & 73.0 & 98.4 & 90.5 & 45.1 & 78.0 & 96.0 & 86.7 & 48.0 & 76.9 \\
   \hline 
   \emph{SM-Prod} & 99.8 & 96.1 & 67.0 & \textbf{87.6} & 99.7 & 97.4 & 54.9 & \textbf{84.0} & 99.7 & 96.7 & 59.1 & \textbf{85.2} \\
   \emph{W. Sum} & 99.7 & 95.8 & 66.6 & 87.4 & 99.7 & 97.2 & 52.8 & 83.2 & 99.7 & 96.4 & 56.9 & 84.3 \\
   \emph{Early} & 76.2  &63.8 & 30.2 & 56.7 & 70.5 & 68.2 & 41.1 & 59.9 & 68.7 & 64.8 & 37.1 & 56.9 \\
  \hline 
\end{tabular}
\end{center}
\end{table}

\subsubsection{State-of-the-art on CASIA-B} \label{subsec:sotaCASIAB}

In Tabs.~\ref{tab:sotaCASIA} and~\ref{tab:sotaCASIACameras}, we compare our results with the \SOTA in CASIA-B under all modalities used before (\textit{Gray} and \textit{OF}) and their fusion. First of all, we would like to remark that our approach uses a resolution of $80\times 60$ while the rest of methods use $320\times 240$. Therefore, our method uses $16$ times less information. %
Regarding \textit{OF} with the 90\textdegree camera, the best results are obtained by PFM~\cite{castro2016ijprai} with a resolution of $320\times 240$. Nevertheless, if we apply it with a resolution of $80\times 60$, its results worsen dramatically and our ResNet-B is able to outperform it in all scenarios. With this modality, our model sets the second best result in the \SOTA (apart from PFM with full resolution). Using all cameras, the best results are achieved by the proposed 3D-CNN although in this case, there are no other approaches to compare with.
If we focus on the visual modality (\textit{Gray} in our case) for the 90\textdegree camera and all cameras, we can see that L-CRF~\cite{chen2018multi} achieves the best results with a resolution of $320\times 240$. However, as this approach is based on dense trajectories, smaller resolutions should produce worse results as we have seen in PFM, which is also based on dense trajectories. For this modality, our best results are obtained by ResNet-B.
Finally, our fusion (softmax product) obtains the best result and it improves our ResNet-B for 90\textdegree camera using \textit{Gray} modality by a $8.3\%$. In the case of using all cameras, our softmax product also improves the results with 2D-CNN by a $8.5\%$.

Comparing our method with~\cite{Wu2016pami}, which is the closest approach to ours as they also employ CNNs, our best average result improves a $16.3\%$ for the 90\textdegree camera and a $14.1\%$ for all cameras with respect to their best average accuracy. Focusing on individual scenarios, they only improve our results in \textit{cl} scenario if we use a single modality, probably due to the use of a gallery-probe scheme, \textit{i.e.}, at test time, they must compare the test sample with all the probe samples to get all distances, then, they select the class of the probe sample with the lowest distance. This method is slower than our approach where we only need to propagate the test sample through the CNN to obtain the class. In addition, if we use our fusion approach, our approach improves them in all scenarios.

\begin{table}[tbh] 
\small
\caption{\textbf{State-of-the-art on CASIA-B, camera $\textbf{90\degree}$}. Percentage of correct rank-1 (R1) recognition for several methods on camera $90\degree$. Bottom rows of each modality correspond to our proposal, where instead of using video frames at $640\times 480$, a resolution of $80\times 60$ is used. Acronyms: `\#subjs' number of subjects used for test; `\#train' number of sequences per person used for training; `\#test' number of sequences per person used for test. Best results are marked in bold. }
\label{tab:sotaCASIA}
\centering
\setlength{\tabcolsep}{0.3em} 
\renewcommand{\arraystretch}{0.9} 
\begin{tabular}{|c|c|c|ccc|ccc|c|}
\hline 
\textit{Modality} & \textit{Input Size} & \textit{Method} & \#subjs & \#train & \#test & \textit{nm} & \textit{bg} & \textit{cl} & \textit{Avg} \\ 
\hline 
\multirow{10}{*}{\rotatebox{90}{Visual Data}} & \multirow{7}{*}{\rotatebox{90}{$320\times 240$}} &GEI~\cite{yu2006casia} & 124 & 4 & 2 & 97.6 & 52.0 & 32.7 &  67.8 \\
&&GEI~\cite{yu2006casia} & 124 & 4 & 2 & 97.6 & 52.0 & 32.7 &  67.8 \\
&&iHMM~\cite{hu2013lbp} & 84 & 5 & 1 & 94.0 & 45.2 & 42.9 &  60.7 \\
&&CGI~\cite{wang2012pami} & 124 & 1 & 1 & 88.1 & 43.7 & 43.0 & 58.3 \\
&& SDCNN~\cite{Alotaibi2015aipr} & 124 & 4 & 2 & 95.6 & - & - & - \\
&& GSP-CTC~\cite{li2018gait} & 124 & 4 & 2 & 99.2 & 77.2 & 73.0 & 83.1 \\
&& L-CRF~\cite{chen2018multi} & 100 & 4 & 2 & 98.6 & 90.2 & 85.8 & \textbf{91.5} \\
\cline{2-10}
&$126\times 126$ & DCNN~\cite{wu2015tmm} & 124 & 4 & 2 & 81.5 & - & - & - \\
\cline{2-10}
&$88\times 128$ & LBCNN~\cite{Wu2016pami} & 50 & 4 & 2 & 91.5 & 63.1 & 54.6 & 69.7 \\ 
\cline{2-10}
&Gray $80\times 60$ & ResNet-B \textbf{(ours)} & 50 & 4 & 2 & 96.0 & 91.0 & 46.0 & 77.7  \\
\hline
\multirow{3}{*}{\rotatebox{90}{OF}} & $320\times 240$ & PFM~\cite{castro2016ijprai} & 124 & 4 & 2 &  100 & 100 & 85.5 & \textbf{95.2} \\
\cline{2-10}
&\multirow{2}{*}{$80\times 60$}&PFM~\cite{castro2016ijprai} & 124 & 4 & 2 & 88.3 & 66.5 & 44.0 & 66.3  \\
& &ResNet-B \textbf{(ours)} & 50 & 4 & 2 & 93.0 & 85.0 & 46.0 & 74.7  \\
\hline
Fusion & $80\times 60$ & ResNet-B-SMP \textbf{(ours)} & 50 & 4 & 2 & 98.0 & 97.0 & 63.0 & \textbf{86.0} \\
\hline 
\end{tabular}
\end{table}

\begin{table}[tbh] 
\small
\caption{\textbf{State-of-the-art on CASIA-B, all cameras}. Percentage of correct rank-1 (R1) recognition for several methods on eleven cameras. Bottom rows of each modality correspond to our proposal, where instead of using video frames at $640\times 480$, a resolution of $80\times 60$ is used. Acronyms: `\#subjs' number of subjects used for test; `\#train' number of sequences per person used for training; `\#test' number of sequences per person used for test. Best results are marked in bold. }
\label{tab:sotaCASIACameras}
\centering
\setlength{\tabcolsep}{0.3em} 
\renewcommand{\arraystretch}{0.9} 
\begin{tabular}{|c|c|c|ccc|ccc|c|}
\hline 
\textit{Modality} & \textit{Input Size} & \textit{Method} & \#subjs & \#train & \#test & \textit{nm} & \textit{bg} & \textit{cl} & \textit{Avg} \\ 
\hline 
\multirow{2}{*}{Visual Data} & $88\times 128$ & LBCNN~\cite{Wu2016pami} & 50 & 4 & 2 & 94.1 & 72.4 & 54.0 & 73.5 \\
\cline{2-10}
&Gray $80\times 60$ & ResNet-B \textbf{(ours)} & 50 & 4 & 2 & 99.5 & 92.3 & 45.6 & 79.1  \\
\hline
OF & $80\times 60$ &3D-CNN \textbf{(ours)} & 50 & 4 & 2 & 98.4 & 90.5 & 45.1 & 78.0  \\
\hline
Fusion & $80\times 60$ & 2D-CNN-SMP \textbf{(ours)} & 50 & 4 & 2 & 99.8 & 96.1 & 67.0 & \textbf{87.6} \\
\hline 
\end{tabular}
\end{table}

\subsection{Released material}\label{subsec:release}
In order to make reproducible the experimental results obtained in this paper, 
the CNN models obtained during the experiments have been publicly released for the research community at the following website:\\
\url{https://github.com/avagait/cnngaitmm}

After the review process, we also plan to release the related source code for reproducing the experiments.

\section{Conclusions} \label{sec:conclu}
We have presented a comparative study of multimodal systems based on CNN architectures for the problem of people identification based on the way the walk (\ie gait recognition). 
The evaluated architectures are able to extract automatically gait signatures from sequences of gray pixels, optical flow and depth maps. Those gait signatures have been tested on the task of people identification, obtaining state-of-the-art results on two challenging datasets, \ie TUM-GAID and CASIA-B, that cover diverse scenarios (\eg people wearing long coats, carrying bags, changing shoes or camera viewpoint changes).

With regard to the type of input modalities, we may conclude that, under similar viewpoints (\eg TUM-GAID) the weakest one is \textit{gray pixels}, as it is highly appearance dependant. 
However, as it could be expected \textit{optical flow} is the one that better encodes body motion. Depth maps work fairly well if changes in appearance are small (\ie \textit{shoes} scenario). In datasets with multiple viewpoints (\eg CASIA-B), \textit{gray pixels} achieve the best results, probably due to \textit{optical flow} produces extremely different vectors depending on the viewpoint so, during training, the optimization process is not able to build a good multiview representation of the subjects.

Regarding the type of architecture, 2D-CNN produces better results in most cases; 3D-CNN is specially useful in scenarios with appearance changes; and, ResNet models are designed to be very deep, therefore, they need huge datasets with high variability between samples to perform well. This has been demonstrated in our experiments where ResNet-A produces worse results than the other two architectures for TUM-GAID (dataset with a single camera viewpoint and few samples) but, on the other hand, ResNet-B produces the best results for CASIA-B (dataset with multiple camera viewpoints and more samples than TUM-GAID).

Finally, the experimental results show that the fusion of multiple modalities allows to boost the recognition accuracy of the system in many cases or at least, it matches the best results achieved by using a single modality. 


As final recommendation and, according to the results obtained, the best models are 3D-CNN and ResNet, being the latter the best option if the dataset contains enough training data. Regarding  fusion methods, the best option is late fusion approaches and, in our case, product of the softmax scores. 

\new{As future work, we plan to study the effect of including new different modalities, such as body silhouettes or human pose. Moreover, we want to study the impact of using temporal components, like LSTM layers, in the performance of the system.}



%
%

\section*{Acknowledgements}
This work has been funded by project TIC-1692 (Junta de Andaluc\'ia).
We gratefully acknowledge the support of NVIDIA Corporation with the donation of the Titan X Pascal GPU used for this research.
Portions of the research in this paper use the CASIA Gait Database collected by Institute of
Automation, Chinese Academy of Sciences.


{
\bibliographystyle{spbasic}
\bibliography{longstrings,local,bibAVA}
}

\end{document}